\documentclass{article}

\usepackage{amsfonts,amssymb,amsmath}
\usepackage{graphicx}
\usepackage{graphics}
\newcommand{\R}{\mathbb{R}}
\newcommand{\Z}{\mathbb{Z}}
\newcommand{\N}{\mathbb{N}}
\newcommand{\Q}{\mathbb{Q}}

\begin{document}

\markboth{Fionn Murtagh}{Hierarchical Clustering in Massive Datasets}

\title{Hierarchical Clustering for Finding Symmetries and Other
Patterns in Massive, High Dimensional Datasets}

\author{Fionn Murtagh (1, 2) and Pedro Contreras (2) \\
(1) Science Foundation Ireland, Wilton Park House, \\
Wilton Place, Dublin 2, Ireland \\ 
and \\
(2) Department of Computer Science \\
Royal Holloway, University of London \\
Egham TW20 0EX, UK \\
\  \\
fmurtagh@acm.org}

\maketitle

\begin{abstract}
Data analysis and data mining are concerned with unsupervised 
pattern finding and structure determination in data sets.  
``Structure'' can be understood as symmetry and a range of 
symmetries are expressed by hierarchy.  Such symmetries directly point 
to invariants, that pinpoint intrinsic properties of the data and of 
the background empirical domain of interest.  We review many 
aspects of hierarchy here, including ultrametric topology, 
generalized ultrametric, linkages with lattices and other discrete
algebraic structures and with p-adic number representations.  By focusing
on symmetries in data we have a powerful means of structuring 
and analyzing massive, high dimensional data stores.  We illustrate 
the powerfulness of hierarchical clustering in case studies in chemistry
and finance, and we provide pointers to other published case studies.
\end{abstract}


\section{Introduction: Hierarchy and Other Symmetries in Data Analysis}

Herbert A.\ Simon, Nobel Laureate in Economics, originator of ``bounded
rationality'' and of ``satisficing'', believed in hierarchy at the basis 
of the human and social sciences, as the following quotation shows:
``... my central theme is that complexity frequently takes the form
of hierarchy and that hierarchic systems have some common properties
independent of their specific content.  Hierarchy, I shall argue, is
one of the central structural schemes that the architect of complexity
uses.'' (\cite{simon}, p. 184.)

Partitioning a set of observations \cite{steinley,steinley2,mirkin} leads to some 
very simple symmetries.  This is one approach to clustering and data mining.
But such  approaches, often based on optimization, 
are not of direct interest to us here.  Instead we will pursue the
theme pointed to by Simon, namely that the notion of 
hierarchy is fundamental for interpreting data and the complex reality 
which the data expresses.   Our work is very different too from the 
marvelous view of the development of mathematical group theory -- but 
viewed in its own right as a complex, evolving system -- presented by 
Foote \cite{foote07}.  

Weyl \cite{weyl} makes the case for the fundamental importance of 
symmetry in science, engineering, architecture, art and other areas.  
As a ``guiding principle'', ``Whenever you have to do with a structure-endowed
entity ... try to determine its group of automorphisms, the group of those
element-wise transformations which leave all structural relations undisturbed.
You can expect to gain a deep insight in the constitution of [the 
structure-endowed entity] in this way.  After that you may start to 
investigate symmetric configurations of elements, i.e.\ configurations which 
are invariant under a certain subgroup of the group of all automorphisms; ...''
(\cite{weyl}, p.\ 144).  

\subsection{About this Article}

In section \ref{sect6}, we describe ultrametric topology as an expression 
of hierarchy.  This provides comprehensive background on the commonly 
used quadratic computational time (i.e., $O(n^2)$, where $n$ is the number of 
observations) agglomerative hierarchical clustering algorithms.   

In section \ref{genum}, we look at the generalized ultrametric 
context.  This is closely linked to analysis based on lattices.  We 
use a case study from chemical database matching to illustrate
algorithms in this area.  

In section \ref{sect3}, p-adic encoding, providing a number 
theory vantage point on ultrametric topology, gives rise to additional
symmetries and ways to capture invariants in data. 

Section \ref{sect4} deals with symmetries that are part and parcel of  
a tree, representing a partial order on data, or equally a set of 
subsets of the data, some of which are embedded.  An application of
such symmetry targets from a dendrogram expressing a hierarchical
embedding is provided through the Haar wavelet transform of a 
dendrogram and wavelet filtering based on the transform.  

Section \ref{sect17} deals with new and recent results relating to the 
remarkable symmetries of massive, and especially high dimensional 
data sets.  An example is discussed of segmenting a financial 
forex (foreign exchange) trading signal.  

\subsection{A Brief Introduction to Hierarchical Clustering}

For the reader new to analysis of data a very short introduction is now
provided on hierarchical clustering.  Along with other families of algorithm,
the objective is automatic classification, for the purposes of data mining,
or knowledge discovery.  Classification, after all, is fundamental in 
human thinking, and machine-based decision making.  But we draw attention 
to the fact that our objective is {\em unsupervised}, as opposed to 
{\em supervised} classification, also known as discriminant analysis or 
(in a general way) machine learning.  So here we are {\em not} concerned
with generalizing the decision making capability of training data, nor are
we concerned with fitting statistical models to data so that these models 
can play a role in generalizing and predicting.  Instead we are concerned
with having ``data speak for themselves''.  That this unsupervised objective
of classifying data (observations, objects, events, phenomena, etc.) 
is a huge task in our society is unquestionably true.  One may think of 
situations when precedents are very limited, for instance.  

Among families of clustering, or unsupervised classification, algorithms,
we can distinguish the following: (i) array permuting and other visualization
approaches; (ii) partitioning to form (discrete or overlapping) clusters 
through optimization, including graph-based approaches; and -- of interest
to us in this article -- (iii) embedded clusters interrelated in a 
tree-based way.  

For the last-mentioned family of algorithm, agglomerative building 
of the hierarchy from consideration of object pairwise distances has 
been the most common approach adopted.  As comprehensive background
texts, see \cite{mirkin0,dubesjain,xu,jaincs}.

\subsection{A Brief Introduction to p-Adic Numbers}

The real number system, and a p-adic number system for given prime,
p, are potentially equally useful alternatives.  
p-Adic numbers were introduced by Kurt Hensel in 1898.  

Whether we deal with Euclidean or with non-Euclidean geometry, 
we are (nearly) always dealing with reals.  But the reals start with the
natural numbers, and from associating observational facts and details 
with such numbers we begin the process of measurement.  From the natural 
numbers, we proceed to the rationals, allowing fractions to be taken 
into consideration.  

The following view of how we do science or carry out other
quantitative study was proposed by Volovich in 1987 \cite{vol87a,vol87b}.
See also the surveys in \cite{PANUAA,freund}.
We can always use rationals to make measurements.  
But they will be approximate, in general.  It is better 
therefore to allow for observables being ``continuous, i.e.\ 
endow them with a topology''.  Therefore  we need a completion of the 
field $\Q$ of rationals.  
To complete the field $\Q$ of rationals, we need Cauchy sequences and 
this requires a norm on $\Q$ (because the Cauchy sequence must converge, 
and a norm is the tool used to show this).  
There is the Archimedean norm such that: 
for any $x, y \in \Q$, with $|x| < |y|$, then there exists
an  
integer $N$ such that $|N x| > |y|$.  For convenience here, we write:
$ |x|_\infty$ for this norm.
So if this completion is Archimedean, then we have $\R = \Q_\infty$, the 
reals.  That is fine if space is taken as commutative and Euclidean.

What of alternatives? Remarkably all norms are known.  
Besides the $\Q_\infty$ norm, we have an infinity of norms, 
$|x|_p$, labeled by 
primes, p.  By Ostrowski's theorem \cite{ostrowski}
these are all the possible norms on $\Q$.  
So we have an unambiguous labeling, via p, of the infinite set of 
non-Archimedean completions of $\Q$ to a field endowed with a 
topology.  

In all cases, we obtain locally compact completions, $\Q_p$, of $\Q$.
They are the fields of p-adic numbers.  All these $\Q_p$ are continua.
Being locally compact, they have additive and multiplicative Haar
measures.  As such we can integrate over them, such as for the 
reals.  

\subsection{Brief Discussion of p-Adic and m-Adic Numbers}
\label{sect23}

We will use p to denote a prime, and m to denote a non-zero positive 
integer. A p-adic number is such that 
any set of p integers which are in distinct residue classes modulo p 
may be used as p-adic digits.  (Cf. remark below, at the end of 
 section \ref{padenc}, quoting from \cite{gouvea}.  It makes the point 
that this opens up a range of alternative notation options in practice.)  
Recall that a ring does not allow division, while a field does.  
m-Adic numbers form a ring; but p-adic numbers form a field.  So 
a priori, 10-adic numbers form a ring.  This provides us with a reason 
for preferring p-adic over m-adic numbers.

We can consider various p-adic expansions:

\begin{enumerate}
\item $\sum_{i=0}^n a_i p^i$, which defines positive integers.
For a p-adic number, we require $a_i \in {0, 1, ... p-1}$.
(In practice: just write the integer in binary form.)
\item $\sum_{i=-\infty}^n a_i p^i$  defines rationals.
\item $\sum_{i=k}^\infty a_i p^i$   where $k$ is an integer, 
not necessarily positive, defines the field $\Q_p$ of p-adic numbers. 
\end{enumerate}

$\Q_p$, the field of p-adic numbers, is (as seen in these definitions) 
the field of p-adic expansions.  

The choice of p is a practical issue.   Indeed, adelic numbers 
use all possible 
values of p (see \cite{brekke} for extensive use and discussion of the 
adelic number framework).  Consider \cite{dragovich,khrenn2}.  DNA 
(desoxyribonucleic acid) 
is encoded using four nucleotides: A, adenine; G, guanine; C, cytosine; and 
T, thymine.  In RNA (ribonucleic acid) T is replaced by U, uracil.  In 
\cite{dragovich} a 5-adic encoding is used, since 5 is a prime and thereby 
offers uniqueness.  In \cite{khrenn2} a 4-adic encoding is used, and a 2-adic
encoding, with the latter based on 2-digit boolean expressions for the four
nucleotides (00, 01, 10, 11).  A default norm is used, based on a 
longest common prefix -- with p-adic digits from the start or left of the 
sequence (see section \ref{sect54} below where this longest common prefix
norm or distance is used and, before that, section \ref{sect333333} where
an example is discussed in detail).  

\section{Ultrametric Topology}
\label{sect6}

In this section we mainly explore symmetries related to: geometric shape; 
matrix structure; and lattice structures.

\subsection{Ultrametric Space for Representing Hierarchy}

\begin{figure}[t]
\begin{center}
\includegraphics[width=8cm]{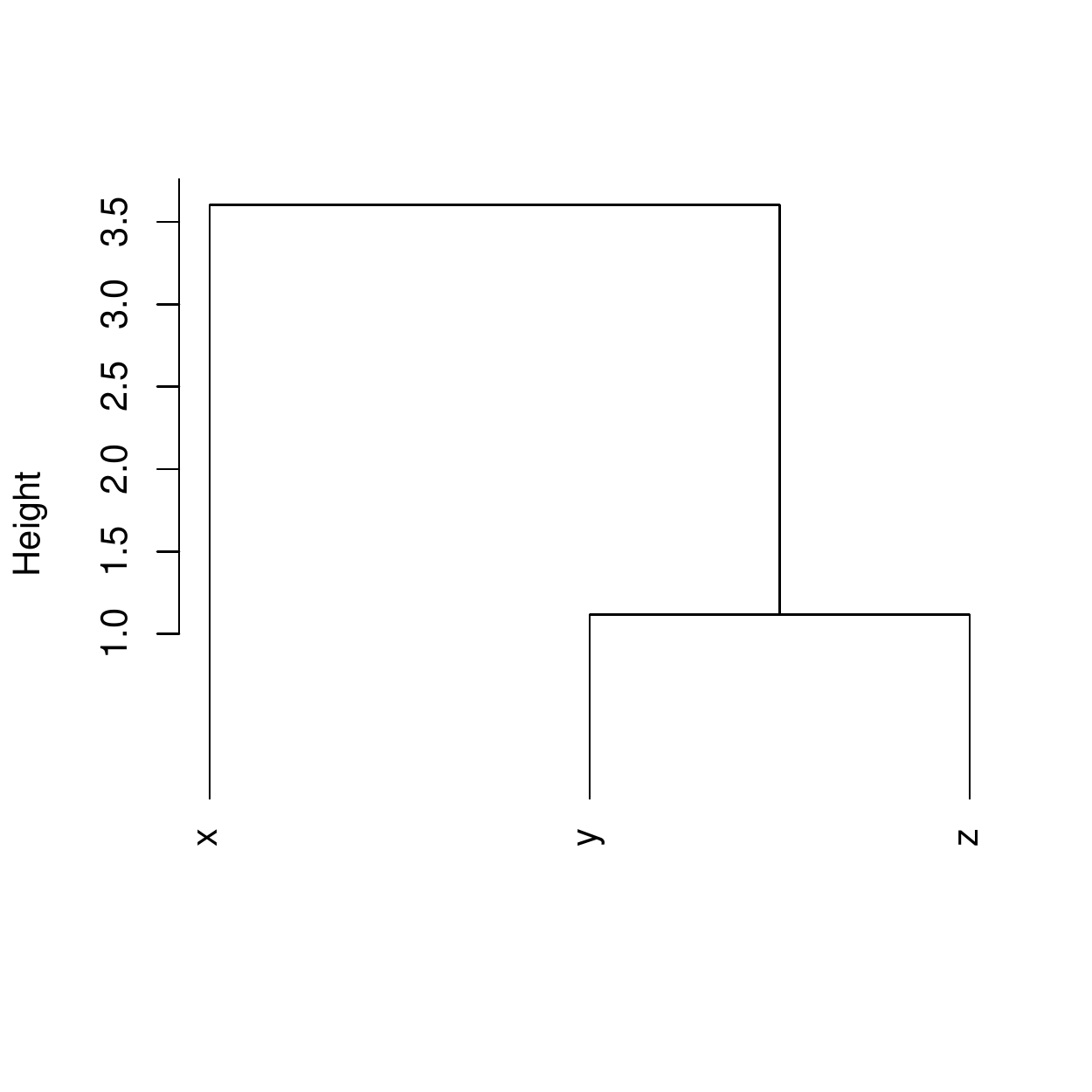}
\end{center}
\caption{The strong triangular inequality defines an ultrametric:
every triplet of points satisfies the relationship:
$d(x,z) \leq \mbox{max} \{ d(x,y), d(y,z) \}$ for distance $d$.
Cf.\ by reading off the hierarchy, how this is verified for all $x, y,      
z$: $d(x,z) = 3.5; d(x,y) = 3.5; d(y,z) = 1.0$.
In addition the symmetry and positive definiteness conditions
hold for any pair of points.}
\label{UMfig0}
\end{figure}

\begin{figure}[t]
\begin{center}
\includegraphics[width=8cm]{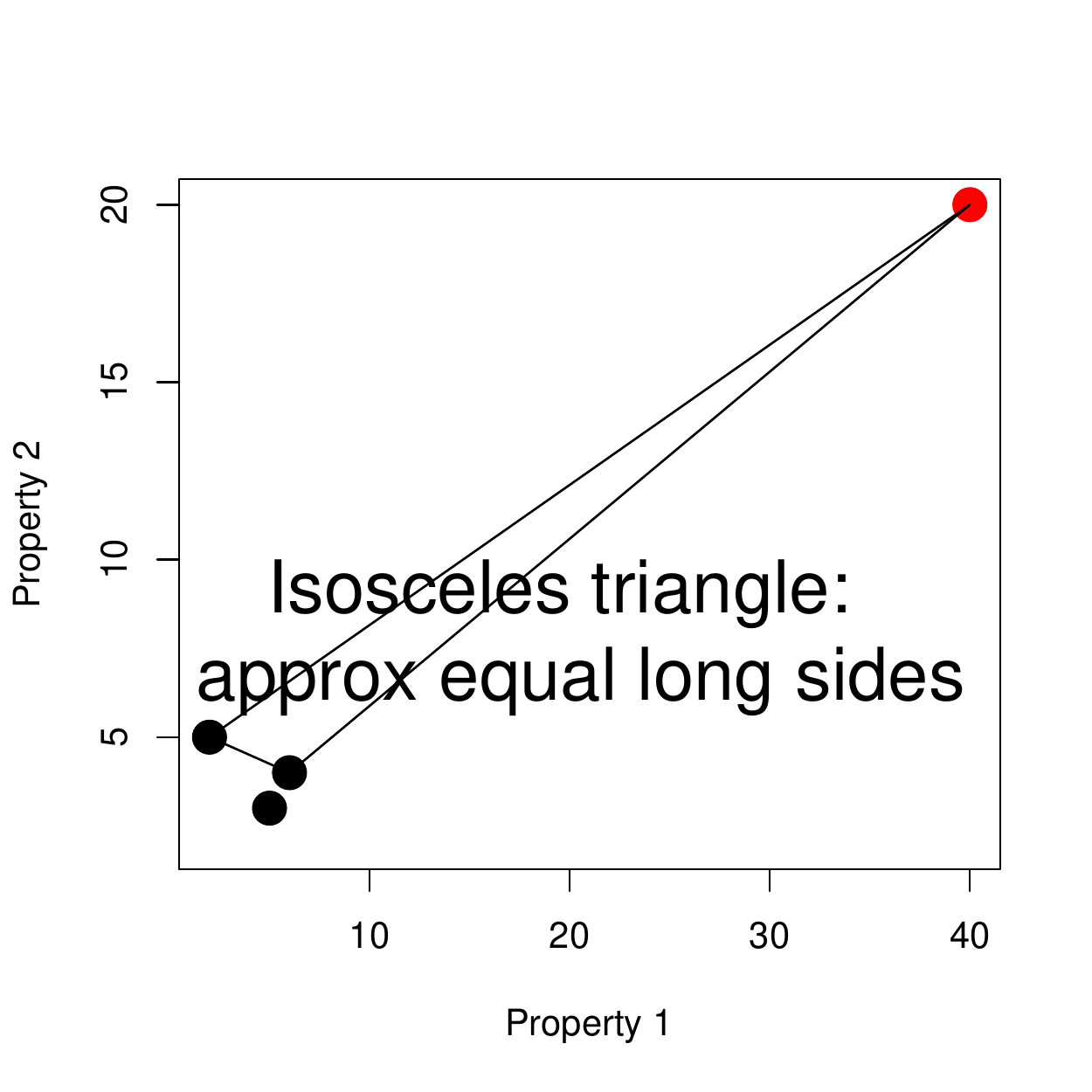}
\end{center}
\caption{How metric data can approximate an ultrametric, or can be 
made to approximate an ultrametric in the case of a stepwise,
agglomerative algorithm.  A ``query'' is on the far right.  While 
we can easily
determine the closest target (among the three objects represented
by the dots on the left), is the closest really that much different
from the alternatives?  This question motivates an ultrametric view 
of the metric relationships shown.}
\label{UMfig000}
\end{figure}

Consider Figures \ref{UMfig0} and \ref{UMfig000}, 
illustrating the ultrametric distance 
and its role in defining a hierarchy.  An early, influential paper 
is Johnson \cite{john} and an important survey is that of 
Rammal et al.\ \cite{rammal86}.  
Discussion of how a hierarchy expresses the semantics of
change and distinction can be found in \cite{murboole}.

The ultrametric topology was introduced by Marc Krasner \cite{kras}, 
the ultrametric inequality having been formulated by Hausdorff in 1934.
 Essential motivation for the study of this area is
provided by \cite{schi} as follows.  Real and complex fields gave rise
to the idea of studying any field $K$ with a complete valuation $| . |$
comparable to the absolute value function.  Such fields satisfy the
``strong triangle inequality'' $| x + y | \leq \mbox{max} ( | x |,
| y | )$.  Given a valued field, defining a totally ordered Abelian 
(i.e.\ commutative) group,
an ultrametric space is induced through $| x - y | = d(x, y)$.
Various terms are used interchangeably for analysis in and
over such fields such as p-adic, ultrametric, non-Archimedean, and isosceles.
The natural geometric ordering of metric valuations is on the real line,
whereas in the ultrametric case the natural ordering is a hierarchical
tree.  

\subsection{Some Geometrical Properties of Ultrametric Spaces}
\label{propum}

We see from the following, based on \cite{lerm} (chapter 0, part IV), 
that an ultrametric space is quite different from a metric one.  In an 
ultrametric space everything ``lives'' on a tree.

In an ultrametric space, all triangles are either isosceles with 
small base, or equilateral.  We have here very clear symmetries of 
shape in an ultrametric topology.   These symmetry ``patterns'' can be 
used to fingerprint data data sets and time series: see 
\cite{murt04,murtaghEPJ} for many examples of this.  

Some further properties that are studied in \cite{lerm} are: 
(i) Every point of a circle in an ultrametric space is a center of the circle.
(ii) In an ultrametric topology, every ball is both open and closed
(termed clopen).
(iii) An ultrametric space is 0-dimensional (see \cite{chak,vanRooij}).
It is clear that an ultrametric topology is very different from our 
intuitive, or Euclidean, notions.  The most important point to keep 
in mind is that in an ultrametric space everything ``lives'' in a 
hierarchy expressed by a tree.

\subsection{Ultrametric Matrices and Their Properties}

For an $n \times n$  matrix of positive reals, symmetric with
respect to the principal diagonal, to be a matrix of distances associated
with an ultrametric distance on $X$, a sufficient and necessary condition
is that a permutation of rows and columns satisfies the following form
of the matrix:

\begin{enumerate}
\item Above the diagonal term, equal to 0, the elements of the same row
are non-decreasing.
\item For every index $k$, if
$$d(k, k+1) = d(k, k+2) =  \dots = d(k, k+ \ell + 1)$$
then
$$d(k+1, j ) \leq d(k,j)  \mbox{  for  } k + 1 < j \leq k + \ell + 1$$
and
$$d(k+1, j) = d(k, j) \mbox{  for  } j > k + \ell + 1$$
Under these circumstances, $\ell \geq 0$ is the length of the section
beginning, beyond the principal diagonal, the interval of columns of
equal terms in row $k$.
\end{enumerate}

\begin{figure}[t]
\begin{center}
\includegraphics[width=9cm]{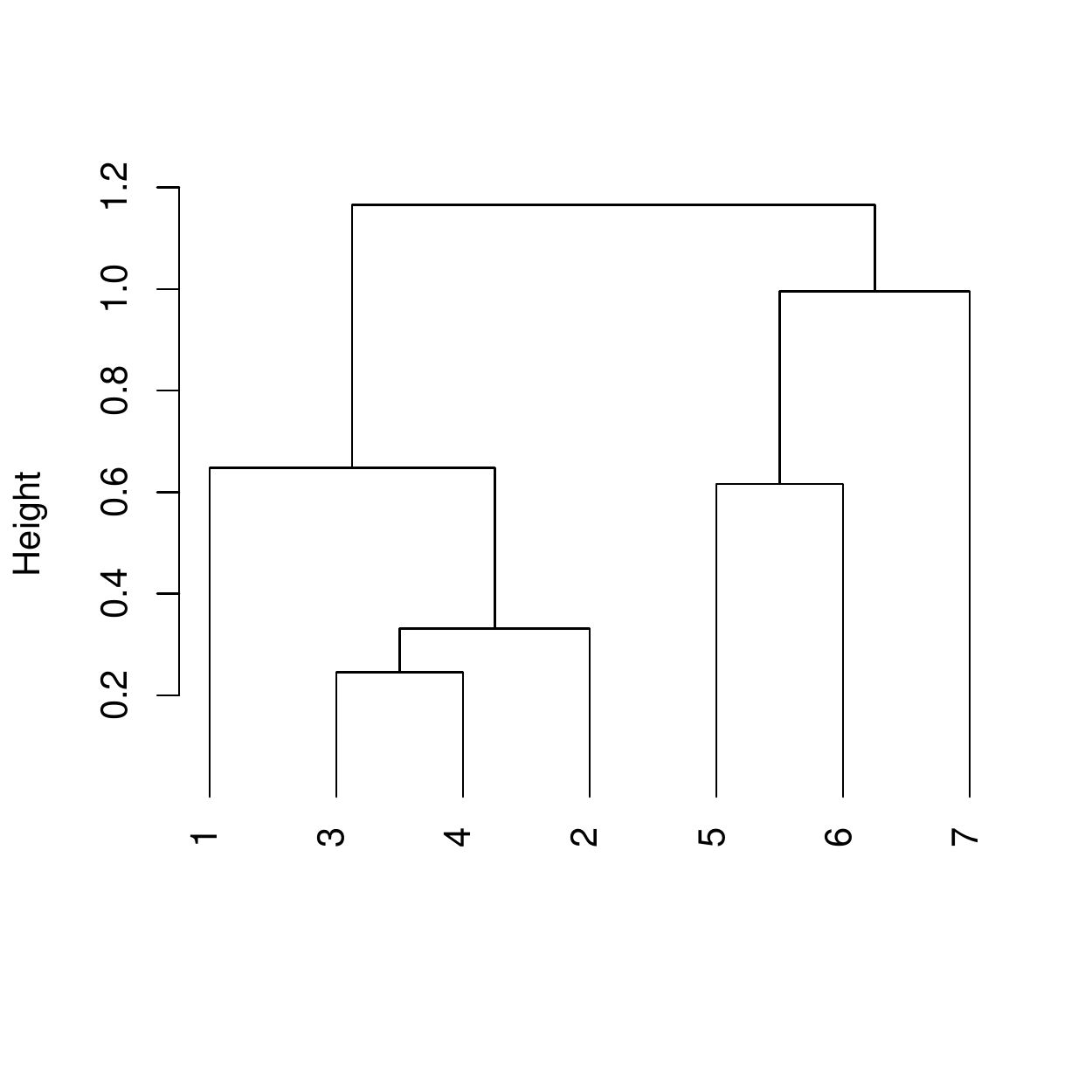}
\end{center}
\caption{Hierarchical clustering of 7 iris flowers using 
data from Table \ref{UMtab1}.  No data normalization was 
used.  The agglomerative clustering criterion was the 
minimum variance or Ward one.}
\label{UMfig1}
\end{figure}

\begin{figure}[t]
\begin{center}
\includegraphics[width=9cm]{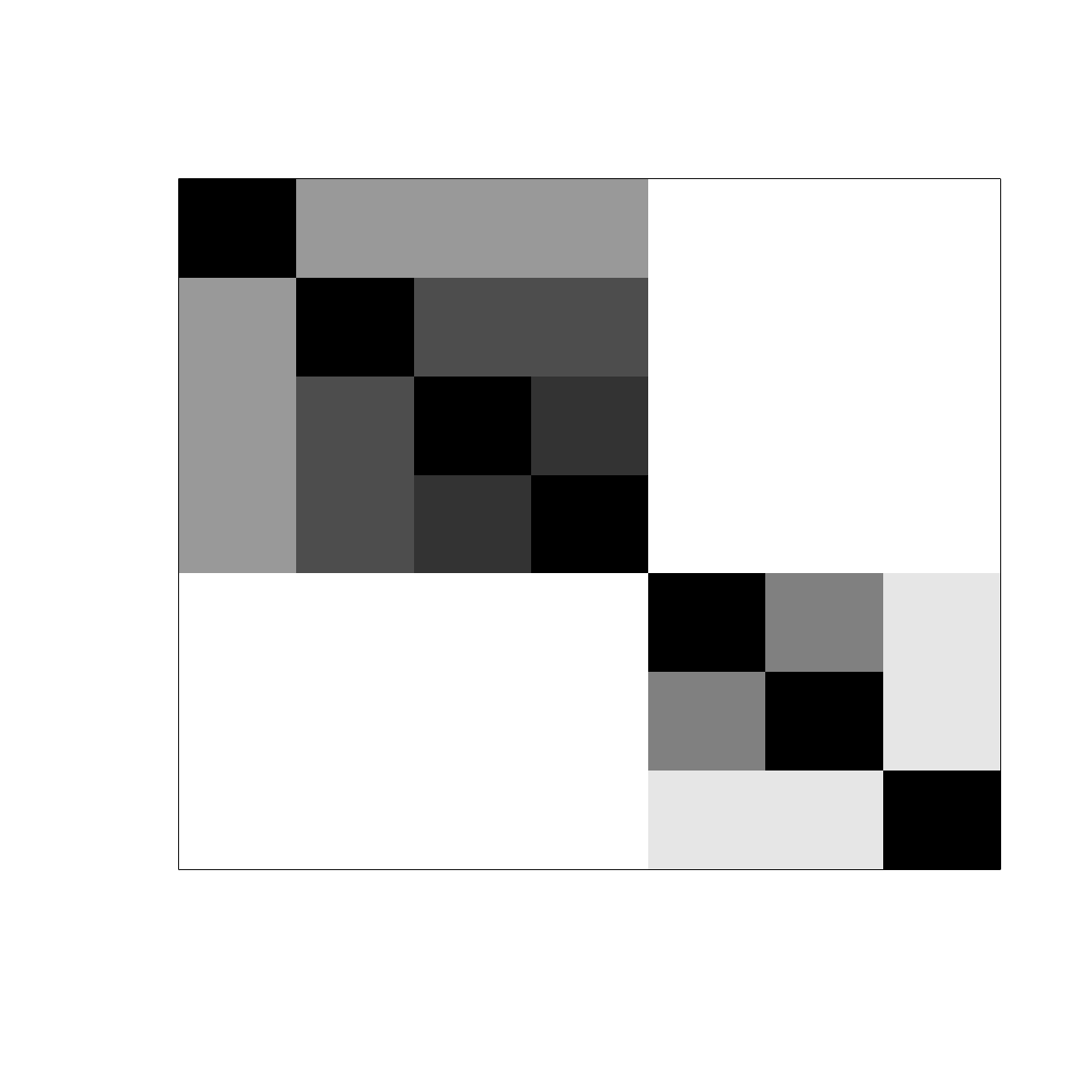}
\end{center}
\caption{A visualization of the ultrametric matrix of 
Table \ref{UMtab2}, where bright or white = highest 
value, and black = lowest value.}
\label{UMfig2}
\end{figure}

To illustrate the ultrametric matrix format, consider the small 
data set shown in Table \ref{UMtab1}.  A dendrogram produced from
this is in Figure \ref{UMfig1}.  The ultrametric matrix that can 
be read off this dendrogram is shown in Table  \ref{UMtab2}.  
Finally a visualization of this matrix, illustrating the ultrametric
matrix properties discussed above, is in Figure \ref{UMfig2}. 

\begin{table}
\begin{center}
\begin{tabular}{|ccccc|} \hline
  & Sepal.Length & Sepal.Width & Petal.Length & Petal.Width \\ \hline
iris1 &  5.1     &    3.5      &    1.4     &    0.2 \\
iris2 & 4.9      &   3.0       &   1.4      &   0.2 \\
iris3 & 4.7      &   3.2       &   1.3      &   0.2 \\
iris4 & 4.6      &   3.1       &   1.5      &   0.2 \\
iris5 & 5.0      &   3.6       &   1.4      &   0.2 \\
iris6 & 5.4      &   3.9       &   1.7      &   0.4 \\
iris7 & 4.6      &   3.4       &   1.4      &   0.3 \\ \hline
\end{tabular}
\end{center}
\caption{Input data: 8 iris flowers characterized by sepal and petal
widths and lengths.  From Fisher's iris data \cite{fisher}.}
\label{UMtab1}
\end{table}

\begin{table}
\begin{tabular}{|llllllll|} \hline
      &  iris1   &   iris2   &   iris3    &  iris4    &  iris5   &   iris6    &  iris7  \\ \hline
iris1 & 0        &  0.6480741 & 0.6480741 & 0.6480741 & 1.1661904 & 1.1661904 & 1.1661904 \\
iris2 & 0.6480741 & 0         & 0.3316625 & 0.3316625 & 1.1661904 & 1.1661904 & 1.1661904 \\
iris3 & 0.6480741 & 0.3316625 & 0         & 0.2449490 & 1.1661904 & 1.1661904 & 1.1661904 \\
iris4 & 0.6480741 & 0.3316625 & 0.2449490 & 0         & 1.1661904 & 1.1661904 & 1.1661904 \\
iris5 & 1.1661904 & 1.1661904 & 1.1661904 & 1.1661904 & 0         & 0.6164414 & 0.9949874 \\
iris6 & 1.1661904 & 1.1661904 & 1.1661904 & 1.1661904 & 0.6164414 & 0         & 0.9949874 \\
iris7 & 1.1661904 & 1.1661904 & 1.1661904 & 1.1661904 & 0.9949874 & 0.9949874 & 0 \\ \hline
\end{tabular}
\caption{Ultrametric matrix derived from the dendrogram in Figure
\ref{UMfig1}.}
\label{UMtab2}
\end{table}

\subsection{Clustering Through Matrix Row and Column Permutation}
\label{sect35}

Figure \ref{UMfig2} shows how an ultrametric distance 
allows a certain structure to be visible (quite possibly, 
in practice, subject to an appropriate row and 
column permuting), in a matrix defined from the set of all 
distances.  For set $X$, then, this matrix expresses the 
distance mapping of the Cartesian product, 
$d: X \times X \longrightarrow \R^+$.  $\R^+$ denotes the 
non-negative reals.  A priori the rows and 
columns of the function of the Cartesian product set $X$ with itself 
could be in any order.  The ultrametric matrix properties establish
what is possible when the distance is an ultrametric one. Because
the matrix (a 2-way data object) involves one {\em mode} (due to 
set $X$ being crossed with itself; as opposed to the 2-mode case where 
an observation set is crossed by an attribute set) 
it is clear that both rows and
columns can be permuted to yield the {\em same} order on $X$. A property of 
the form of the matrix is that small values are at or near the principal
diagonal.   

A generalization opens up for this sort of clustering by visualization 
scheme.  Firstly, we can directly apply row and column permuting to
2-mode data, i.e.\ to the rows and columns of a matrix crossing 
indices $I$ by attributes $J$, $a : I \times J \longrightarrow \R$.
A matrix of values, $a(i,j)$, is furnished by the function $a$ acting on 
the sets $I$ and $J$.  
Here, each such term is real-valued.  We can also generalize the 
principle of permuting such that small values are on or near the principal
diagonal to instead allow similar values to be near one another, and thereby
to facilitate visualization.  
An optimized way to do this was pursued in \cite{cormick,march}.
Comprehensive surveys of clustering algorithms in this area, including 
objective functions, visualization schemes, optimization approaches, 
presence of constraints, and applications, can be found in \cite{mech,mad}.
See too \cite{deutsch,murt85}.

For all these approaches, underpinning them are row and column
permutations, that can be expressed in terms of the permutation group, $S_n$, 
on $n$ elements.  

\subsection{Other Miscellaneous Symmetries}

As examples of various other local symmetries worthy of 
consideration in data sets consider subsets of 
data comprising clusters, and reciprocal nearest neighbor pairs.  

Given an observation set, $X$, we define dissimilarities as the mapping 
$d : X \times X \longrightarrow \R^+$.
A dissimilarity is a positive, definite, symmetric measure (i.e., $d(x, y)
\geq 0; d(x,y) = 0 \mbox{ if } x = y; d(x,y) = d(y,x)$).  If in addition
the triangular inequality is satisfied (i.e., $d(x,y) \leq d(x,z) + d(z,y),
\forall x, y, z \in X$) then the dissimilarity is a distance.  

If $X$ is endowed with a metric, then this metric is 
mapped onto an ultrametric.  In practice, there is no need for $X$ to be
endowed with a metric.  Instead a dissimilarity is satisfactory.

A hierarchy, $H$,
is defined as a binary, rooted, node-ranked tree, also
termed a dendrogram \cite{benz,john,lerm,murt85}. 
A hierarchy defines a set of embedded subsets of a given set of objects
$X$, indexed by the set $I$.  That is to say, object $i$ in the 
object set $X$ is denoted $x_i$, and $i \in I$.  
These subsets are {\em totally ordered} by an index function $\nu$, which is a
stronger condition than the {\em partial order} 
required by the subset relation.  The index function $\nu$ is represented
by the ordinate in Figure \ref{UMfig1} (the ``height'' or ``level'').  
A bijection exists between a hierarchy and an ultrametric space.

Often in this article we will refer interchangeably to the object set,
$X$, and the associated set of indices, $I$.  

Usually a constructive approach is used to induce $H$ on a set $I$.  
The most efficient algorithms are based on nearest neighbor chains, which 
by definition end in a pair of agglomerable reciprocal nearest neighbors.
Further information can be found in \cite{murt83,murt84,murt85,murt92}.




\section{Generalized Ultrametric}
\label{genum}

In this subsection, we consider an ultrametric defined on the power set 
or join semilattice.  Comprehensive background on ordered sets and 
lattices can be found in \cite{davey}.  
A review of generalized distances and ultrametrics 
can be found in \cite{sedacj}.

\subsection{Link with Formal Concept Analysis}

Typically hierarchical clustering is based on a distance (which can be
relaxed often to a dissimilarity, not respecting the triangular inequality,
and {\em mutatis mutandis} to a similarity), defined on all pairs of the object
set: $d: X \times X \rightarrow \R^+$.  I.e., a distance is  a positive
real value.  Usually we require that a distance cannot be 0-valued unless
the objects are identical.  That is the traditional approach.  

A different form of 
ultrametrization is achieved from a dissimilarity defined on the power set
of attributes characterizing the observations (objects, individuals, etc.)
$X$.  Here we have: $d : X \times X \longrightarrow 2^J$, where $J$ 
indexes  the attribute (variables, characteristics, properties, etc.) set.

This gives rise to a different notion of distance, that maps pairs of objects
onto elements of a join semilattice.  The latter can represent all subsets
of the attribute set, $J$.  That is to say, it can represent the power set,
commonly denoted $2^J$, of $J$.  

As an example, consider, say, $n = 5$ objects characterized by 3 boolean
(presence/absence) attributes, shown in Figure \ref{figfca} (top).
Define dissimilarity between a pair of objects in this table as
a {\em set} of 3 components, corresponding to the 3 attributes, such that
if both components are 0, we have 1; if either component is 1 and the
other 0, we have 1; and if both components are 1 we get 0.  This is the
simple matching coefficient \cite{jan1}.  We could
use, e.g., Euclidean distance for each of the values sought; but we prefer
to treat 0 values in both components as signaling  a 1 contribution.  We get
then $d(a,b) = 1, 1, 0$ which we will call \verb+d1,d2+.  Then, 
$d(a,c) = 0, 1, 0$ which we will call \verb+d2+.  Etc.  
With the latter we create lattice nodes as shown in the middle 
part of Figure \ref{figfca}.

\begin{figure}
\begin{center}
\begin{tabular}{cccc}
   &  $v_1$  &   $v_2$  & $v_3$  \\
a  &    1    &    0     &   1    \\
b  &    0    &    1     &   1    \\
c  &    1    &    0     &   1    \\
e  &    1    &    0     &   0    \\
f  &    0    &    0     &   1    \\
\end{tabular}
\end{center}
\begin{verbatim}

Potential lattice vertices      Lattice vertices found       Level

       d1,d2,d3                        d1,d2,d3                3
                                         /  \
                                        /    \
  d1,d2   d2,d3   d1,d3            d1,d2     d2,d3             2
                                        \    /
                                         \  /
   d1      d2      d3                     d2                   1

\end{verbatim}

   The set d1,d2,d3 corresponds to:     $d(b,e)$ and $d(e,f)$

   The subset d1,d2 corresponds to:     $d(a,b), d(a,f), d(b,c),
                                        d(b,f),$ and $d(c,f)$

   The subset d2,d3 corresponds to:     $d(a,e)$ and $d(c,e)$

   The subset d2 corresponds to:         $d(a,c)$

\medskip

   Clusters defined by all pairwise linkage at level $\leq  2$:

$   a, b, c, f$

$   a, c, e$

\medskip

   Clusters defined by all pairwise linkage at level $\leq 3$:

$   a, b, c, e, f$

\caption{Top: example data set consisting of 5 objects,
characterized by 3 boolean attributes.
Then: 
lattice corresponding to this data and its interpretation.}
\label{figfca}
\end{figure}

In Formal Concept Analysis \cite{davey,ganter}, 
it is the lattice itself which is
of primary interest.  In \cite{jan1} there is discussion of, and a range 
of examples on,  the close
relationship between the traditional hierarchical cluster analysis
based on $d: I \times I \rightarrow \R^+$, and hierarchical cluster
analysis ``based on abstract posets'' (a poset is a partially ordered
set), based on $d: I \times I \rightarrow 2^J$.  The latter, leading to
clustering based on dissimilarities, was developed initially
 in \cite{jan0}.  

\subsection{Applications of Generalized Ultrametrics}
\label{sect332}

As noted in the previous subsection, the usual 
ultrametric is an ultrametric distance, i.e.\ for a set I,
$d: I \times I \longrightarrow \R^+$. The generalized
ultrametric is also consistent with this definition, where the 
range is a subset of the power set: 
$d: I \times I \longrightarrow \Gamma$, 
where $\Gamma$ is a partially ordered set.  In other words, the 
{\em generalized} ultrametric distance is a set.  Some areas of 
application of generalized ultrametrics will now be discussed.  

In the theory of reasoning, a  monotonic operator is rigorous application 
of a succession of  conditionals (sometimes called consequence 
relations).  However negation or multiple valued logic (i.e.\ 
encompassing intermediate truth and falsehood) require support for 
non-monotonic reasoning.  

Thus \cite{seda}: 
``Once one introduces negation 
... then certain of the important
operators are not monotonic (and therefore not continuous), and in 
consequence the Knaster-Tarski theorem [i.e.\ for fixed points; 
see \cite{davey}] is 
no longer applicable to them.  Various ways have been proposed to 
overcome this problem.  One such [approach is to use] syntactic conditions on 
programs ... Another is to consider different operators ... The 
third main solution is to introduce techniques from topology and 
analysis to augment arguments based on order ... [the latter include:]
methods based on metrics ... on quasi-metrics ... and finally ...
on ultrametric spaces.''

The convergence to fixed points that are based on a generalized 
ultrametric system is precisely the study of 
spherically complete systems and expansive automorphisms
discussed in section \ref{dilat} below.  As expansive automorphisms we see here
again an example of symmetry at work.  

\subsection{Example of Application: Chemical Database Matching}
\label{sect333333}

In the 1990s, the Ward minimum variance hierarchical clustering method
became the method of choice in the
chemoinformatics community due to its hierarchical nature and the quality of
the clusters produced. Unfortunately the method reached its limits once the
pharmaceutical companies tried processing datasets of more than 500,000
compounds due to:
the $O(n^2)$ processing requirements of the reciprocal nearest
neighbor  algorithm;
the requirement to hold all chemical structure ``fingerprints'' in
memory to enable random access; and the requirement that parallel
implementation use a shared-memory architecture.  Let us look at
an alternative hierarchical clustering
algorithm that bypasses these computational difficulties. 

A direct application of generalized ultrametrics 
to data mining is the following.  
The potentially huge advantage of the generalized ultrametric is that 
it allows a hierarchy to be read directly off the $I \times J$ input 
data, and bypasses the $O(n^2)$ consideration of all pairwise distances 
in agglomerative hierarchical clustering.  In \cite{sisc} we study 
application to chemoinformatics.  Proximity and best match finding is 
an essential operation in this field.  Typically we have one million 
chemicals upwards, characterized by an approximate 1000-valued attribute
encoding.  

Consider first our need to normalize the data.  We divide
each boolean (presence/absence) value by its
corresponding column sum.  

We can consider the hierarchical
cluster analysis from abstract posets as based on $d: I \times I              
\rightarrow \R^{|J|}$.
In \cite{jan1}, the median of the $|J|$ distance values is used, as input
to a traditional
hierarchical clustering, with alternative schemes discussed.  See also
\cite{jan0} for an early elaboration of this approach.

Let us now proceed to take a particular approach to this, which has
very convincing computational benefits.  

\subsubsection{Ultrametrization through Baire Space Embedding: Notation}
\label{sect331}

A Baire space \cite{levy} consists of countably infinite sequences with
a metric defined in terms of the longest common prefix: the longer the
common prefix, the closer a pair of sequences.  The Baire metric, and 
simultaneously ultrametric, will be defined in definition 
\ref{baire} in the next subsection.  What is of interest to us
here is this longest common prefix metric, which additionally is 
an ultrametric.  The longest common prefixes at issue here
are those of precision of any value (i.e., $x_{ij}$, for chemical
compound $i$, and chemical structure code $j$).  Consider two such values,
$x_{ij}$ and $y_{ij}$, which, when the context easily allows it, we will
call $x$ and $y$.  Each are of some precision, and we take the integer
$|K|$ to be the maximum precision.  We pad a value with 0s if necessary,
so that all values are of the same precision.  Finally, we will assume
for convenience that each value $\in [0, 1)$  and this can be arranged by
normalization.

\subsubsection{The Case of One Attribute}
\label{sect332332}

Thus we consider ordered sets
$x_{k}$ and $y_{k}$ for  $k \in K$.  In line with
our notation, we can write $x_K$ and $y_K$ for these
numbers, with the set $K$ now ordered.  (So, $k = 1$ is the first decimal
place of precision; $k = 2$ is the second decimal place; $\dots$ ;
$k = |K|$ is the $|K|$th decimal place.)
The cardinality of the set $K$ is the precision with which
a number, $x_K$, is measured.  Without loss of generality, through
normalization,   we will take
all $x_K, y_K \leq 1$.    We will also consider decimal numbers,
only, in this article (hence $x_k \in \{ 0, 1, 2, \dots , 9 \}$
for all numbers
$x$, and for all digits $k$), again with no loss of generality to non-decimal
number representations.

Consider as examples $x_K = 0.478$; and
$y_K = 0.472$.  In these cases, $| K | = 3$.  For $k = 1$, we find $x_k       
= y_k = 4$.  For $k = 2, x_k = y_k$.  But for $k = 3, x_k \neq y_k$.

We now introduce the following distance:

\begin{equation}
d_B(x_K, y_K) =
\begin{cases}
1                   & \mbox{if } x_1 \neq y_1  \\
\mbox{inf } 2^{-n}  & x_n = y_n  \ \ \  1 \leq n \leq |K|
\end{cases}
\label{baire}
\end{equation}


So for $x_K = 0.478$ and $y_K = 0.472$ we have $d_B(x_K, y_K) = 
2^{-2} = 0.25$.  

The Baire distance is used in denotational semantics where one
considers $x_K$ and $y_K$ as words (of equal length, in the finite case),
and then this distance is defined from a common
$n$-length prefix, or left substring, in the two words.  For a set of
words, a prefix tree can be built to expedite word matching, and the Baire
distance derived from this tree.

We have $1 \geq d_B(x_K, y_K) \geq 2^{-|K|}$.  Identical $x_K$ and
$y_K$ have Baire distance equal to $2^{-|K|}$.  The Baire distance
is a 1-bounded ultrametric.

The Baire ultrametric defines a hierarchy, which can be expressed as a
multiway tree, on a set of numbers, $x_{IK}$.  So the number $x_{iK}$,
indexed by $i$, $i \in I$,  is of precision $|K|$.  It is actually
simple to determine this hierarchy.
The partition at level $k = 1$ has
clusters defined as all those numbers indexed by $i$ that share the same
1st digit.
The partition at level $k = 2$ has
clusters defined as all those numbers indexed by $i$ that share the same
2nd digit; and so on, until we reach $k = |K|$.
A strictly finer, or identical, partition is to be found at each
successive level (since once a pair of numbers becomes dissimilar, $d_B > 0$,
this non-zero distance cannot be reversed).  Identical numbers at
level $k = 1$ have distance $ \leq 2^{-1} = 0.5$.
 Identical numbers at level
$k = 2$ have distance $\leq 2^{-2} = 0.25$.
 Identical numbers at level
$k = 3$ have distance $\leq 2^{-3} = 0.125$; and so on, to level $k = |K|$,
when distance $= 2^{-|K|}$.

\subsubsection{Analysis: Baire Ultrametrization from Numerical Precision}
\label{sect5}

In this section we use (i) a random projection of vectors into a 1-dimensional
space (so each chemical structure is mapped onto a scalar value, by design
$\geq 0$ and $\leq 1$) followed by (ii) implicit use of a prefix tree
constructed on the digits of the set of scalar values.  First we will look
at this procedure.  Then we will return to discuss its properties.

We seek all $i, i^\prime$ such that:
\begin{enumerate}
\item for all $j \in J$,
\item $x_{ijK} = x_{i^\prime jK}$
\item to fixed precision $K$
\end{enumerate}
Recall that $K$ is an ordered set.  We impose a user specified upper
limit on precision, $|K|$.

Now rather than $|J|$ separate tests for equality (point 1 above), a
{\em sufficient condition} is that $\sum_j w_j x_{ijK} = \sum_j w_j           
x_{i^\prime j K}$ for a set of weights $w_j$.
What helps in making
this sufficient condition for equality work well in practice is that
many of the $x_{iJK}$ values are 0: cf.\ the approximate 8\% matrix
occupancy rate that holds here.
We experimented with such
possibilities as $w_j = j$ (i.e., $\{ 1, 2, \dots , |J| \}$ and
$w_j = |J| + 1 - j$ (i.e., $\{ |J|, |J| - 1, \dots , 3, 2, 1 \}$.  A
first principal component would allow for the definition of the
least squares optimal linear fit of the projections.  The best choice of
$w_j$ values we found for uniformly distributed values in $(0,1)$:
for each $j$, $w_j \sim U(0,1)$.

\begin{table}
\begin{center}
\begin{tabular}{|c|r|}\hline
Sig. dig. $c$ &  No. clusters  \\ \hline
           &                \\
  4        &    6591        \\
  4        &    6507        \\
  4        &    5735        \\
           &                \\
  3        &    6481        \\
  3        &    6402        \\
  3        &    5360        \\
           &                \\
  2        &    2519        \\
  2        &    2576        \\
  2        &    2135        \\
           &                \\
  1        &     138        \\
  1        &     148        \\
  1        &     167        \\
           &                \\ \hline
\end{tabular}
\end{center}
\caption{Results for the three different data sets, each consisting of
7500 chemicals,  are shown in immediate
succession.  The number of significant decimal digits is 4 (more precise,
and hence more different clusters found), 3, 2, and 1 (lowest precision
in terms of significant digits).}
\label{tab2}
\end{table}

Table \ref{tab2} shows, in immediate succession,
results for three data sets.  The normalizing
column sums were calculated and applied independently to each of the
three data sets.  Insofar as $x_J$ is directly proportional,
 whether calculated on 7500 chemical structures or 1.2 million, leads to a
constant of proportionality, only, between the two cases.
As noted, a 
random projection was used.  Finally, identical projected values were
read off, to determine clusters.

\subsubsection{Discussion: Random Projection and Hashing}
\label{sect334}

Random projection
is the finding of a low dimensional embedding of a point set -- dimension
equals 1, or a line or axis, in this work  -- such that the
distortion of any pair of points is bounded by a function of the lower
dimensionality \cite{vempala}.  There is a burgeoning literature in this
area, e.g.\ \cite{dutta}.
While random projection {\em per se} will not  guarantee
a bijection of best match in original and in lower dimensional spaces,
our use of projection here is effectively a hashing method (\cite{cox}
uses MD5 for nearest neighbor search), in
order to deliberately find hash collisions -- thereby providing
a {\em sufficient} condition for the mapped vectors to be identical.

Collision of identically valued vectors is guaranteed, but what of
collision of non-identically valued vectors, which we want to avoid?

To prove such a result may require an assumption of what distribution
our original data follow.  A general class is referred to as a stable
distribution \cite{indyk}: this is a distribution such that a limited
number of
weighted sums of the  variables is also itself of the same distribution.
Examples include both Gaussian and long-tailed or power law distributions.

Interestingly, however, very high dimensional (or equivalently, very low
sample size or low $n$) data sets, by
virtue of high relative dimensionality alone, have points mostly lying
at the vertices of a regular simplex or polygon \cite{murt04,hall}.  This
intriguing aspect is one reason, perhaps, why we have found random projection
to work well.  Another reason is the following: if we work on normalized
data, then the values on any two attributes $j$ will be small.  Hence
$x_j$ and $x'_j$ are small.  Now if the random weight for this attribute is
$w_j$, then the random projections are, respectively, $\sum_j w_j x_j$ and
$\sum_j w_j x'_j$.  But these terms are dominated by the random weights.
We can expect near equal $x_j$ and $x'_j$ terms, for all $j$, to be mapped
onto fairly close resultant scalar values.

Further work is required to confirm these hypotheses, viz.,
that high dimensional
data may be highly ``regular'' or ``structured'' in such a way;
and that, as a consequence, hashing is particularly well-behaved
in the sense of non-identical vectors being nearly always
collision-free.  There is further discussion in \cite{pedrophd}.  

We remark that a  prefix tree, or trie, is well-known in the 
searching and sorting
literature \cite{gusfield},
and is used to expedite the finding of longest common
prefixes.  At level one, nodes are associated with  the first digit.
At level two, nodes are associated with the second digit, and so on
through deeper levels of the tree.

\subsubsection{Simple Clustering Hierarchy from the Baire                        
Space Embedding}
\label{simpleclust}

The Baire ultrametrization induces a (fairly flat) multiway tree
on the given data set.

Consider a partition yielded by identity (over all the attribute set)
at a given precision level.  Then for precision levels $k_1, k_2, k_3,        
\dots$ we have, at each, a partition, such that all member clusters are
ordered by reverse embedding (or set inclusion): $q_{(1)} \supseteq q_{(2)}   
\supseteq q_{(3)} \supseteq \dots$.  Call each such sequence of embeddings a
chain.  The entire data set is covered by a set of such chains.
This sequence
of partitions is ordered by set inclusion.

The computational time complexity is as follows.  Let the
number of chemicals be denoted $n = |I|$; the number of attributes is
$|J|$; and the total number of digits precision is $|K|$.  Consider a
particular number of digits precision, $k_0$, where $ 1 \leq k_0 \leq         
|K|$.  Then the random projection takes $ n \cdot k_0 \cdot |J|$ operations.
A sort follows, requiring $O(n \log n)$ operations.  Then clusters are
read off with $O(n)$ operations.  Overall, the computational effort is
bounded by $ c_1 \cdot  |I| \cdot |J| \cdot |K| + c_2 \cdot |I| \cdot \log |I|
+ c_3 |I|$ (where $c_1, c_2, c_3$ are constants), which is equal to
$O(|I| \log |I|)$ or $O(n \log n)$.

Further evaluation and a number of further case studies are 
covered in \cite{pedrophd}.  

\section{Hierarchy in a p-Adic Number System} 
\label{sect3}

A dendrogram is widely used in hierarchical, agglomerative clustering,
and is induced from observed data.  In this article, one of our 
important goals is to 
show how it lays bare many diverse symmetries in the observed phenomenon 
represented by the data.  By expressing a dendrogram in p-adic terms,
we open up a wide range of possibilities for seeing symmetries and
attendant invariants.  

\subsection{p-Adic Encoding of a Dendrogram}
\label{padenc}

We will introduce now the one-to-one mapping of clusters (including
singletons) in a dendrogram
$H$ into a set of p-adically expressed integers (a forteriori,
rationals, or $\Q_p$).
The field of p-adic numbers is the most important example of
ultrametric spaces.
Addition and multiplication of p-adic integers, $\Z_p$ (cf.\ expression 
in subsection \ref{sect23}), are well-defined.  Inverses
exist and no zero-divisors exist.

A terminal-to-root traversal in a dendrogram or binary rooted tree
is defined as follows.  We use
the path $x \subset q \subset q' \subset q'' \subset \dots q_{n-1}$, where
$x$ is a given object specifying a given terminal, and $q, q', q'', \dots$
are the embedded classes along this path, specifying nodes in the
dendrogram.  The root node is specified by the class $q_{n-1}$
comprising all objects.

A terminal-to-root traversal is the shortest path between the given terminal
node and the root node,
assuming we preclude repeated traversal (backtrack)
of the same path between any two nodes.

\begin{figure}
\centering
\includegraphics[width=14cm]{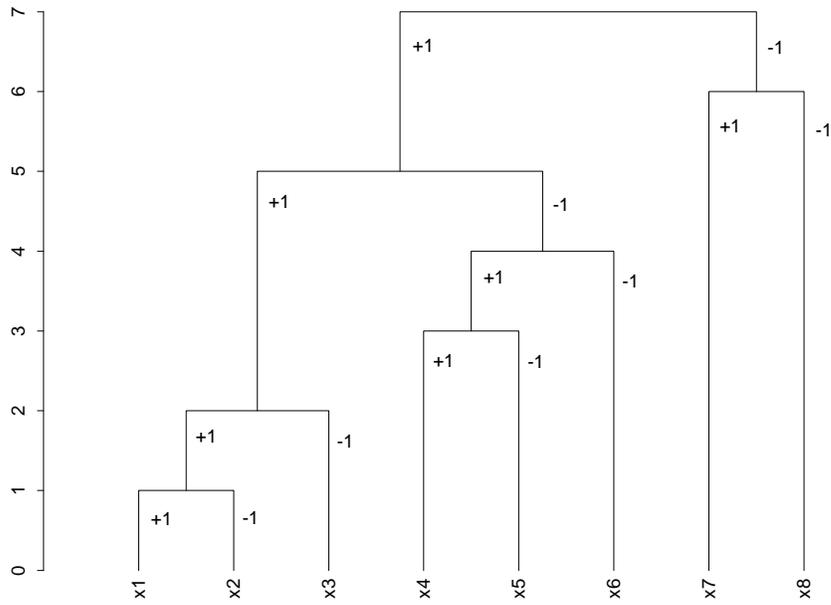}
\caption{Labeled, ranked dendrogram on 8 terminal nodes,
$x_1, x_2, \dots , x_8$.  Branches are labeled
$+1$ and $-1$. Clusters are: $q_1 = \{ x_1, x_2 \},
q_2 = \{ x_1, x_2, x_3 \}, q_3 = \{ x_4, x_5 \}, q_4 =  \{ x_4, x_5,
x_6 \}, q_5 = \{ x_1, x_2, x_3, x_4, x_5, x_6 \},
q_6 = \{ x_7, x_8 \}, q_7 = \{ x_1, x_2, \dots , x_7, x_8 \}$.}
\label{fig2}
\end{figure}

By means of terminal-to-root traversals, we define the following
p-adic encoding of terminal nodes, and hence objects, in Figure \ref{fig2}.

\begin{eqnarray}
x_1: & + 1 \cdot p^1 + 1 \cdot p^2 + 1 \cdot p^5 + 1 \cdot p^7  \\ \nonumber
x_2: & - 1 \cdot p^1 + 1 \cdot p^2 + 1 \cdot p^5 + 1 \cdot p^7  \\ \nonumber
x_3: & - 1 \cdot p^2 + 1 \cdot p^5 + 1 \cdot p^7  \\ \nonumber
x_4: & + 1 \cdot p^3 + 1 \cdot p^4 - 1 \cdot p^5 + 1 \cdot p^7   \\ \nonumber
x_5: & - 1 \cdot p^3 + 1 \cdot p^4 - 1 \cdot p^5 + 1 \cdot p^7   \\ \nonumber
x_6: & - 1 \cdot p^4 - 1 \cdot p^5 + 1 \cdot p^7  \\ \nonumber
x_7: & + 1 \cdot p^6 - 1 \cdot p^7   \\ \nonumber
x_8: & - 1 \cdot p^6 - 1 \cdot p^7
\label{eqn00}
\end{eqnarray}

If we choose $p = 2$ the resulting decimal equivalents could be the same:
cf.\ contributions based on $+1 \cdot p^1$ and $-1 \cdot p^1 + 1 \cdot p^2$.
Given that the coefficients of the $p^j$
terms ($1 \leq j \leq 7$) are in the set $\{ -1, 0, +1 \}$
(implying for $x_1$ the additional terms: $+ 0 \cdot p^3 + 0 \cdot p^4
+ 0 \cdot p^6$),
the coding based on $p = 3$ is required to avoid ambiguity among
decimal equivalents.

A few general remarks on this encoding follow.  For the labeled ranked
binary trees that we are considering  (for discussion of combinatorial 
properties based on labeled, ranked and binary trees, see 
\cite{ref11a}), we require the labels $+1$ and $-1$
for the two branches at any node.  Of course we could interchange
these labels,
and have these $+1$ and $-1$ labels reversed at any node.
By doing so we will have different p-adic codes for the objects, $x_i$.

The following properties hold: (i) {\em Unique encoding:} the
decimal codes for
each $x_i$ (lexicographically ordered) are unique for $p \geq 3$;
and (ii) {\em Reversibility:} the dendrogram can be uniquely reconstructed
from any such set of unique codes.

The p-adic encoding defined for any object set
can be expressed as follows for any object $x$
associated with a terminal node:

\begin{equation}
x = \sum_{j=1}^{n-1} c_j p^j  \mbox{    where } c_j \in \{ -1, 0, +1 \}
\label{eqn1}
\end{equation}

In greater detail we have:

\begin{equation}
x_i = \sum_{j=1}^{n-1} c_{ij} p^j  \mbox{    where } c_{ij}
\in \{ -1, 0, +1 \}
\label{eqn1b}
\end{equation}

Here $j$ is the level or rank (root: $n-1$; terminal: 1), and $i$ is an
object index.

In our example we have used: $c_j = +1$ for a left branch
(in the sense of Figure \ref{fig2}), $= -1$ for a right
branch, and $= 0$ when the node is not on the path from that particular
terminal to the root.

A matrix form of this encoding is as follows, where
$\{ \cdot \}^t$ denotes the transpose of the vector.

Let $\mathbf{x}$ be the column vector $\{ x_1 \ x_2 \ \dots x_n \}^t$.

Let $\mathbf{p}$ be the column vector $\{ p^1 \ p^2 \ \dots p^{n-1} \}^t$.

Define a characteristic matrix $C$ of the branching codes, $+1$ and $-1$,
and an absent or non-existent branching given by $0$,
as a
set of values $c_{ij}$ where $i \in I$, the indices of the object set; and
$j \in \{ 1, 2, \dots , n-1 \}$, the indices of the dendrogram levels
or nodes ordered
increasingly.  For Figure \ref{fig2} we therefore have:

\begin{equation}
C = \{ c_{ij} \} =
\left(
\begin{array}{rrrrrrr}
1  & 1 & 0 &  0 & 1 &  0 & 1 \\
-1 & 1 & 0 &  0 & 1 & 0 & 1 \\
0  & -1 & 0 &  0 & 1 & 0 & 1 \\
0  &  0 & 1 &  1 & -1 & 0 & 1 \\
0  & 0  & -1 & 1 & -1 & 0 & 1 \\
0  & 0 & 0  & -1 & -1 & 0 & 1 \\
0  & 0 & 0  & 0  & 0 & 1 & -1 \\
0  & 0 & 0  & 0  & 0 & -1 & -1
\end{array}
\right)
\label{eqn2}
\end{equation}

For given level $j$, $\forall i$, the absolute values $| c_{ij} |$
give the membership function either by node, $j$, which is therefore
read off
columnwise; or by object index, $i$, which is therefore read off rowwise.

The matrix form of the p-adic encoding used in equations (\ref{eqn1}) or 
(\ref{eqn1b}) is:

\begin{equation}
\mathbf{x} = C \mathbf{p}
\label{eqn3}
\end{equation}

Here, {\bf x} is the decimal encoding, $C$ is the matrix with dendrogram
branching codes (cf.\ example shown in expression (\ref{eqn2})), 
and {\bf p} is the vector of powers of a fixed integer
(usually, more restrictively, fixed prime) $p$.

The tree encoding exemplified in Figure \ref{fig2}, and defined with
 coefficients
in equations (\ref{eqn1}) or (\ref{eqn1b}),
(\ref{eqn2}) or (\ref{eqn3}), with labels  $+ 1 $ and
$ - 1$ was required (as opposed to the choice of 0 and 1, 
which might have been our first thought)
to fully cater for the ranked nodes (i.e.\ the
total order, as opposed to a partial order, on the nodes).

We can
consider the objects that we are dealing with to have equivalent integer
values.  To show that, all we must do is work out decimal equivalents of
the p-adic expressions used above for $x_1, x_2, \dots $.  As noted
in \cite{gouvea}, we have equivalence between: a p-adic number; a
p-adic expansion; and an element of $\Z_p$ (the p-adic integers).
The coefficients used to specify a p-adic number, 
\cite{gouvea} notes
(p.\ 69), ``must be taken in a set of representatives of the class modulo
$p$.  The numbers between 0 and $p-1$ are only the most obvious choice for
these representatives.  There are situations, however, where other choices
are expedient.''

We note that the matrix $C$ is used in \cite{heiser1d}.  A somewhat 
trivial view 
of how ``hierarchical trees can be perfectly scaled in one dimension'' 
(the title and theme of \cite{heiser1d}) is that p-adic numbering is
feasible, and hence a one dimensional representation of terminal nodes
is easily arranged through expressing each p-adic number with a real 
number equivalent.  

\subsection{p-Adic Distance on a Dendrogram}
\label{sect54}

We will now induce a metric topology on the p-adically encoded dendrogram,
$H$. It leads to various symmetries
relative to identical norms, for instance, or identical tree distances.

We use the following longest common subsequence, starting at the root: 
we look for the term $p^r$ in the p-adic
codes of the two objects, where $r$ is the lowest level such that the
values of the coefficients of $p^r$ are equal.

Let us look at the set of p-adic codes for $x_1, x_2, \dots $ above
(Figure \ref{fig2} and relations \ref{eqn00}), to
give some examples of this.

\medskip

For $x_1$ and $x_2$, we find the term we are looking for to be $p^1$, and
so $r = 1$.

For $x_1$ and $x_5$, we find the term we are looking for to be $p^5$, and
so $r = 5$.

For $x_5$ and $x_8$, we find the term we are looking for to be $p^7$, and
so $r = 7$.

\medskip

Having found the value $r$, the distance is defined as $p^{-r}$
\cite{benz,gouvea}.

This longest common prefix metric
is also known as the Baire distance, and has been discussed 
in section \ref{sect333333}.
In topology the Baire metric is defined on
infinite strings \cite{levy}. It is more than just a distance:
it is an ultrametric bounded from above by 1, and
its {\em infimum} is 0 which is relevant for very long sequences, or
in the limit for infinite-length sequences.  The use of this Baire metric
is pursued in \cite{sisc} based on random projections \cite{vempala}, and
providing  computational benefits over the
classical $O(n^2)$ hierarchical clustering based on all pairwise distances.

The longest common prefix metric leads directly to
a {\em p-adic hierarchical classification} (cf.\ \cite{bradley}).
This is a special case of
the ``fast'' hierarchical clustering discussed in section \ref{sect332}.

Compared to the longest common prefix metric,  there are other
related forms of metric, and simultaneously ultrametric.
In \cite{gajic}, the metric
is defined via the integer part of a real number.
In \cite{benz}, for integers $x, y$ we have:
$d(x,y) = 2^{-\mbox{order}_p(x - y)}$
where $p$ is prime, and order$_p(i)$ is the exponent (non-negative integer) of
$p$ in the prime decomposition of an integer.
Furthermore let
$S(x)$ be a series: $S(x) = \sum_{i \in \N}  a_i x^i$.
($\N$ are the natural numbers.)
The order of $S(i)$ is the rank of its first non-zero term:
order$(S) = \inf \{ i : i \in \N; a_i \neq 0 \}$.
(The series that is all zero is of order infinity.)
Then the ultrametric similarity between series is:
$d(S, S') = 2^{- \mbox{order}(S - S') }$.

\subsection{Scale-Related Symmetry}
\label{dilat}

Scale-related symmetry is very important in practice.  In this 
subsection we introduce an operator that provides this symmetry.  We also 
term it a dilation operator, because of its role in the wavelet transform 
on trees (see section \ref{sect53} below, and 
\cite{murhaar} for discussion and examples).  This 
operator is p-adic multiplication by $1/p$.  

Consider the set of objects 
$\{ x_i | i \in I \}$ with its p-adic coding considered
above.  Take $ p = 2$.  (Non-uniqueness of corresponding decimal codes is
not of concern to us now, and taking this value for $p$ is without any
loss of generality.)
Multiplication of
$x_1 = + 1 \cdot 2^1 + 1 \cdot 2^2 + 1 \cdot 2^5 + 1 \cdot 2^7 $ by
$1/p = 1/2$ gives: $  + 1 \cdot 2^1 + 1 \cdot 2^4 + 1 \cdot 2^6$.  Each
level has decreased by one, and the lowest level has been lost.
Subject to the lowest level of the tree being lost, the form of the tree
remains the same.  By carrying
out the multiplication-by-$1/p$ operation on all objects, it is seen that
the effect is to rise in the hierarchy by one level.

Let us call product with $1/p$ the operator $A$.  The effect of losing the
bottom level of the dendrogram means that either (i) each cluster (possibly
singleton) remains the same; or (ii) two clusters are merged.  Therefore
the application of $A$ to all $q$ implies a subset relationship between
the set of clusters $\{ q \}$ and the result of applying $A$, $\{ A q \}$.

Repeated application of the operator $A$ gives $A q$, $A^2 q$,
$A^3 q$, $\dots$.  Starting with any singleton, $i \in I$, this gives
a path from the terminal to the root node in the tree.  Each such
path ends with the null element, which we define to be the p-adic 
encoding corresponding to the root node of the tree.
Therefore the intersection of the paths equals the
null element.   

Benedetto and Benedetto \cite{benben,ben}  discuss $A$ as an expansive
automorphism of $I$, i.e.\ form-preserving, and locally expansive.
Some implications \cite{benben}  of  
the expansive automorphism follow.
For any $q$, let us
take $q, A q, A^2 q, \dots$ as a sequence of open subgroups of
$I$, with $q \subset A q \subset A^2 q \subset \dots$, and $I =
\bigcup \{ q, A q, A^2 q, \dots \} $.  This is termed an inductive sequence
of $I$, and $I$ itself is the inductive limit 
(\cite{reiter}, p.\ 131).  

Each path defined by application of the expansive automorphism
defines a spherically complete system \cite{schi,gajic,vanRooij},
which is a
formalization of well-defined subset embeddedness.
Such a methodological framework finds application in multi-valued
and non-monotonic reasoning, as noted in section \ref{sect332}.  

\section{Tree Symmetries through the Wreath Product Group}
\label{sect4}

In this section the wreath product group, used up to now in 
the literature
as a framework for tree structuring of image or other signal 
data, is here used on a 2-way tree or dendrogram data structure.
An example of wreath product invariance is provided by the 
wavelet transform {\em of} such a tree.

\subsection{Wreath Product Group 
Corresponding to a Hierarchical Clustering}
\label{wreath}

A dendrogram like that shown in Figure \ref{fig2} is invariant 
as a representation or structuring of a data set relative
to rotation (alternatively, here: permutation) of left and right 
child nodes.  
These rotation (or permutation) symmetries are defined 
by the wreath product group (see \cite{foote1,foote2,foote3} for 
an introduction and applications in signal and image processing), 
and can be used with any m-ary tree, 
although we will treat the binary or 2-way case here.

For the group actions, with respect to which we will seek
invariance, we consider independent cyclic shifts of the
subnodes of a given node (hence, at each level).  Equivalently
these actions are adjacency preserving permutations of
subnodes of a given node (i.e., for given $q$, 
with $q = q' \cup q''$, the permutations
of $\{  q', q'' \}$).  
We have therefore cyclic
group actions at each node, where the cyclic group is of order
2.

The symmetries of $H$ are given by structured permutations of
the terminals.  The terminals will be denoted here by
Term $H$. The full group of symmetries is summarized
by the following generative algorithm:

\begin{enumerate}
\item For level  $l = n - 1$ down to 1 do:
\item Selected node, $\nu \longleftarrow $ node at level $l$.
\item And permute subnodes of $\nu$.
\end{enumerate}

Subnode $\nu$ is the root of subtree $H_\nu$.  We denote $H_{n-1}$ simply by
$H$.  For a subnode $\nu'$ undergoing a relocation action in step 3, the
internal structure of subtree $H_{\nu'}$ is not altered.

The algorithm described defines the automorphism group which is a
wreath product of the symmetric group.  Denote the permutation at level
$\nu$ by $P_\nu$.  Then the automorphism group is given by:
$$G = P_{n-1} \ \mathrm{wr} \ P_{n-2} \ \mathrm{wr} \ \dots \ \mathrm{wr} \ P_2
\  \mathrm{wr} \  P_1$$
where wr denotes the wreath product.

\subsection{Wreath Product Invariance}

Call Term $H_\nu$ the terminals that descend from the node at level $\nu$.
So these are the terminals of the subtree $H_\nu$ with its root node at
level $\nu$.  We can alternatively call Term $H_\nu$ the cluster associated
with level $\nu$.

We will now look at shift invariance under the group action.  This amounts
to the requirement for a constant function defined on Term $H_\nu, \forall
\nu$.  A convenient way to do this is to define such a function on the set
Term $H_\nu$ via the root node alone, $\nu$.  By definition then we have a
constant function on the set Term $H_\nu$.

Let us call $V_\nu$ a space of functions that are constant on Term $H_\nu$.
That is to say, the functions are constant in clusters that are defined by
the subset of $n$ objects. 
Possibilities for $V_\nu$ that were considered in \cite{murhaar} are:

\begin{enumerate}
\item Basis  vector with $| \mathrm{Term } H_{n-1} |$ 
components, with 0 values
except for value 1 for component $i$.
\item Set (of cardinality $n = | \mathrm{Term } H_{n-1} |$) of $m$-dimensional
observation vectors.
\end{enumerate}

Consider the resolution scheme arising from moving from \\
Term $ H_{\nu'}$, Term $ H_{\nu''} \}$ to
Term  $H_\nu $.  From the hierarchical clustering point of view it is
clear what this represents, simply, an agglomeration of two clusters
called Term $H_{\nu'}$ and Term $H_{\nu''}$, replacing them with a new
cluster, Term $H_\nu$.

Let the spaces of functions that are constant on subsets
corresponding to the two cluster agglomerands be denoted $V_{\nu'}$ and
$V_{\nu''}$.  These two clusters are disjoint initially, which motivates
us taking the two spaces as a couple: $(V_{\nu'}, V_{\nu''})$.

\subsection{Example of Wreath Product Invariance: Haar Wavelet Transform
of a Dendrogram}
\label{sect53}

Let us exemplify a case that satisfies all that has been
defined in the context of the wreath product invariance that we are
targeting.  It is the  algorithm discussed in depth in
\cite{murhaar}. 
Take the constant function from $V_{\nu'}$ to be $f_{\nu'}$.
Take the constant function from $V_{\nu''}$ to be $f_{\nu''}$.
Then define the constant function, the {\em scaling function},
 in $V_{\nu}$ to be
$(f_{\nu'} + f_{\nu''})/2$.  Next define the zero mean function,
$(w_{\nu'} + w_{\nu''})/2 = 0$, the {\em wavelet function}, as follows:

$$w_{\nu'} = (f_{\nu'} + f_{\nu''})/2 - f_{\nu'}$$
in the support interval of $V_{\nu'}$, i.e.\ Term $H_{\nu'}$, and
$$w_{\nu''} = (f_{\nu'} + f_{\nu''})/2 - f_{\nu''}$$
in the support interval of $V_{\nu''}$, i.e.\ Term $H_{\nu''}$.

Since $w_{\nu'} = - w_{\nu''}$ we have the zero mean requirement.

We now illustrate the Haar wavelet transform of a dendrogram with a 
case study.  

The discrete wavelet transform is a decomposition of data into
spatial and frequency components.  In terms of a dendrogram these
components are with respect to, respectively, within and between clusters of
successive partitions.  We show how this works taking the data
of Table \ref{table5}.

\begin{table}
\begin{center}
\begin{tabular}{|rrrrr|} \hline
    &   Sepal.L   &  Sepal.W    &  Petal.L  &   Petal.W \\ \hline
1   &       5.1   &      3.5    &      1.4  &       0.2 \\
2   &       4.9   &      3.0    &      1.4  &       0.2 \\
3   &       4.7   &      3.2    &      1.3  &       0.2 \\
4   &       4.6   &      3.1    &      1.5  &       0.2 \\
5   &       5.0   &      3.6    &      1.4  &       0.2 \\
6   &       5.4   &      3.9    &      1.7  &       0.4 \\
7   &       4.6   &      3.4    &      1.4  &       0.3 \\
8   &       5.0   &      3.4    &      1.5  &       0.2 \\ \hline
\end{tabular}
\end{center}
\caption{First 8 observations of Fisher's iris data.  L and W
refer to length and width.}
\label{table5}
\end{table}

\begin{figure*}
\begin{center}
\includegraphics[width=14cm]{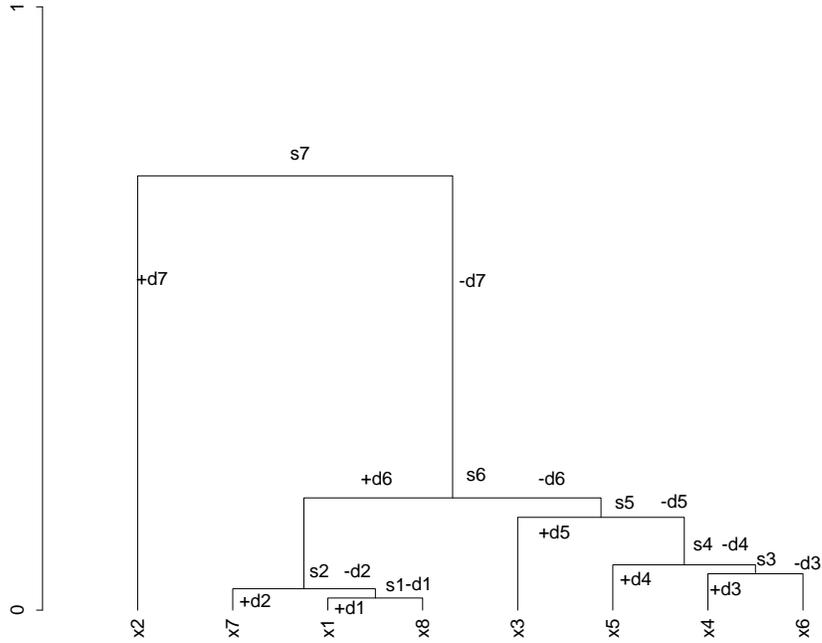}
\end{center}
\caption{Dendrogram on 8 terminal nodes constructed from first 8 values
of Fisher iris data.  (Median agglomerative method used in this case.)
Detail or wavelet coefficients are denoted by $d$, and data smooths are
denoted by $s$.  The observation vectors are denoted by $x$ and are
associated with the terminal nodes.
Each {\em signal smooth}, $s$, is a vector.  The (positive
or negative) {\em detail signals}, $d$, are also vectors.  All these
vectors are of the same dimensionality.}
\label{fig555}
\end{figure*}

The hierarchy built on the 8 observations of Table \ref{table5}
is shown in Figure \ref{fig555}.  Here we note the associations of 
irises 1 through 8 as, respectively: $x_1, x_3, x_4, x_6, x_8, x_2,
x_5, x_7$.  

Something more is shown in Figure \ref{fig555}, namely the
detail signals (denoted $\pm d$) and overall smooth (denoted $s$),
which are determined in carrying out the wavelet transform,
the so-called forward transform.

The inverse transform is then determined from Figure \ref{fig555}
in the following way.  Consider the observation vector $x_2$.  Then
this vector is reconstructed exactly by reading the tree from the
root: $s_7 + d_7 = x_2$.  Similarly a path from root to terminal is
used to reconstruct any other observation.  If $x_2$ is a vector
of dimensionality $m$, then so also are $s_7$ and $d_7$, as well
as all other detail signals.

\begin{table*}
\begin{center}
\begin{tabular}{|rrrrrrrrr|} \hline
        &     s7  &     d7    &    d6  &   d5  &   d4   &  d3  &  d2 & d1 \\
\hline
Sepal.L & 5.146875 & 0.253125 & 0.13125 & 0.1375 &$-0.025$ & 0.05 & $-0.025$ &
0.05 \\
Sepal.W & 3.603125 & 0.296875 & 0.16875 & $-0.1375$ & 0.125 & 0.05 & $-0.075$ &
$-0.05$ \\
Petal.L & 1.562500 & 0.137500 & 0.02500 & 0.0000 & 0.000 & $-0.10$ & 0.050
 & 0.00 \\
Petal.W & 0.306250 & 0.093750 & $-0.01250$ & $-0.0250$ & 0.050 & 0.00 & 0.000
 & 0.00 \\ \hline
\end{tabular}
\end{center}
\caption{The hierarchical Haar wavelet transform resulting from use of the
first 8 observations of Fisher's iris data shown in Table \ref{table5}.
Wavelet coefficient levels are denoted
d1 through d7, and the continuum or smooth component is denoted s7.}
\label{table6}
\end{table*}

This procedure is the same as the Haar wavelet transform, only
applied to the dendrogram and using the input data.

This wavelet transform for the data
in Table \ref{table5}, based on the ``key'' or intermediary
hierarchy of Figure \ref{fig555},  is shown in Table \ref{table6}.

Wavelet regression entails setting small and hence unimportant
detail coefficients to 0 before applying the inverse wavelet transform.
More discussion can be found in \cite{murhaar}.

Early work on p-adic and ultrametric wavelets can be found in 
Kozyrev \cite{kozyrev02,kozyrev07}.  While we have treated the case
of the wavelet transform on a particular graph, a tree, recent 
applications of wavelets to general graphs are in \cite{nason}
and, by representing the graph as a matrix, in \cite{berry}.

\section{Remarkable Symmetries in Very High Dimensional Spaces}
\label{sect17}

In the work of \cite{rammal85,rammal86} it was shown how as ambient dimensionality increased 
distances became more and more ultrametric.  That is to say, a hierarchical embedding becomes
more and more immediate and direct as dimensionality increases.  A better way of 
quantifying this phenomenon was developed in \cite{murt04}.   What this means is that 
there is inherent hierarchical structure in high dimensional data spaces.  

It was shown experimentally in \cite{rammal85,rammal86,murt04} how 
points in high dimensional spaces become increasingly equidistant 
with increase in 
dimensionality.  Both \cite{hall} and \cite{donoho} study Gaussian 
clouds in very 
high dimensions.  The latter finds that ``not only are the points 
[of a Gaussian 
cloud in very high dimensional space] on the convex hull, but all 
reasonable-sized subsets span faces of the convex hull.  This is 
wildly different
than the behavior that would be expected by traditional low-dimensional 
thinking''.

That very simple structures come about in very high dimensions is not as 
trivial as it might appear at first sight.  Firstly, even very simple structures 
(hence with many symmetries) can be used to support fast and perhaps even constant 
time worst case proximity search \cite{murt04}.  Secondly, as shown in 
the machine learning framework by \cite{hall}, 
there are important implications ensuing from the simple high dimensional structures.
Thirdly, \cite{murt08} shows that very high dimensional clustered data contain 
symmetries that in fact can be exploited to ``read off'' 
the clusters in a computationally efficient way.  Fourthly, following 
\cite{delon}, what we might want to look for in contexts of 
considerable symmetry are the ``impurities'' or small irregularities 
that detract from the overall dominant picture.  

See Table \ref{tabunifgauss} exemplifying the change of topological
properties as ambient dimensionality increases.  It behoves us to 
exploit the symmetries that arise when we have to process 
very high dimenionsal data. 

\begin{table}
\begin{center}
\begin{tabular}{lllll} \hline
No. points &    Dimen. &  Isosc. &   Equil. &    UM   \\ \hline
           &           &         &          &         \\
Uniform    &           &         &          &         \\
           &           &         &          &         \\
100        &    20     &    0.10 &    0.03  &    0.13 \\
100        &    200    &    0.16 &    0.20  &    0.36 \\
100        &    2000   &    0.01 &    0.83  &    0.84 \\
100        &    20000  &    0    &    0.94  &    0.94 \\
           &           &         &          &         \\
Hypercube  &           &         &          &         \\
           &           &         &          &         \\
100        &    20     &    0.14 &   0.02   &    0.16 \\
100        &    200    &    0.16 &   0.21   &    0.36 \\
100        &    2000   &    0.01 &   0.86   &    0.87 \\
100        &    20000  &    0    &   0.96   &    0.96 \\
           &           &         &          &         \\
Gaussian   &           &         &          &         \\
           &           &         &          &         \\
100        &    20     &    0.12 &    0.01  &    0.13 \\
100        &    200    &    0.23 &    0.14  &    0.36 \\
100        &    2000   &    0.04 &    0.77  &    0.80 \\
100        &    20000  &    0    &    0.98  &    0.98 \\ \hline
\end{tabular}
\end{center}
\caption{Typical results, based on 300 sampled triangles from triplets of
points.  For uniform, the data are generated on $[0, 1]^m$; hypercube vertices
are in $\{ 0, 1\}^m$,
and for Gaussian on each dimension, the data are
of mean 0, and variance 1.  Dimen.\ is the ambient
dimensionality.  Isosc.\ is the number of isosceles triangles with
small base, as a
proportion of all triangles sampled.  Equil.\ is the number of equilateral
triangles as a proportion of triangles sampled.  UM is the proportion of
ultrametricity-respecting triangles (= 1 for all ultrametric).}
\label{tabunifgauss}
\end{table}

\subsection{Application to Very High Frequency Data Analysis: 
Segmenting a Financial Signal}

\begin{figure*}
\includegraphics[width=16cm]{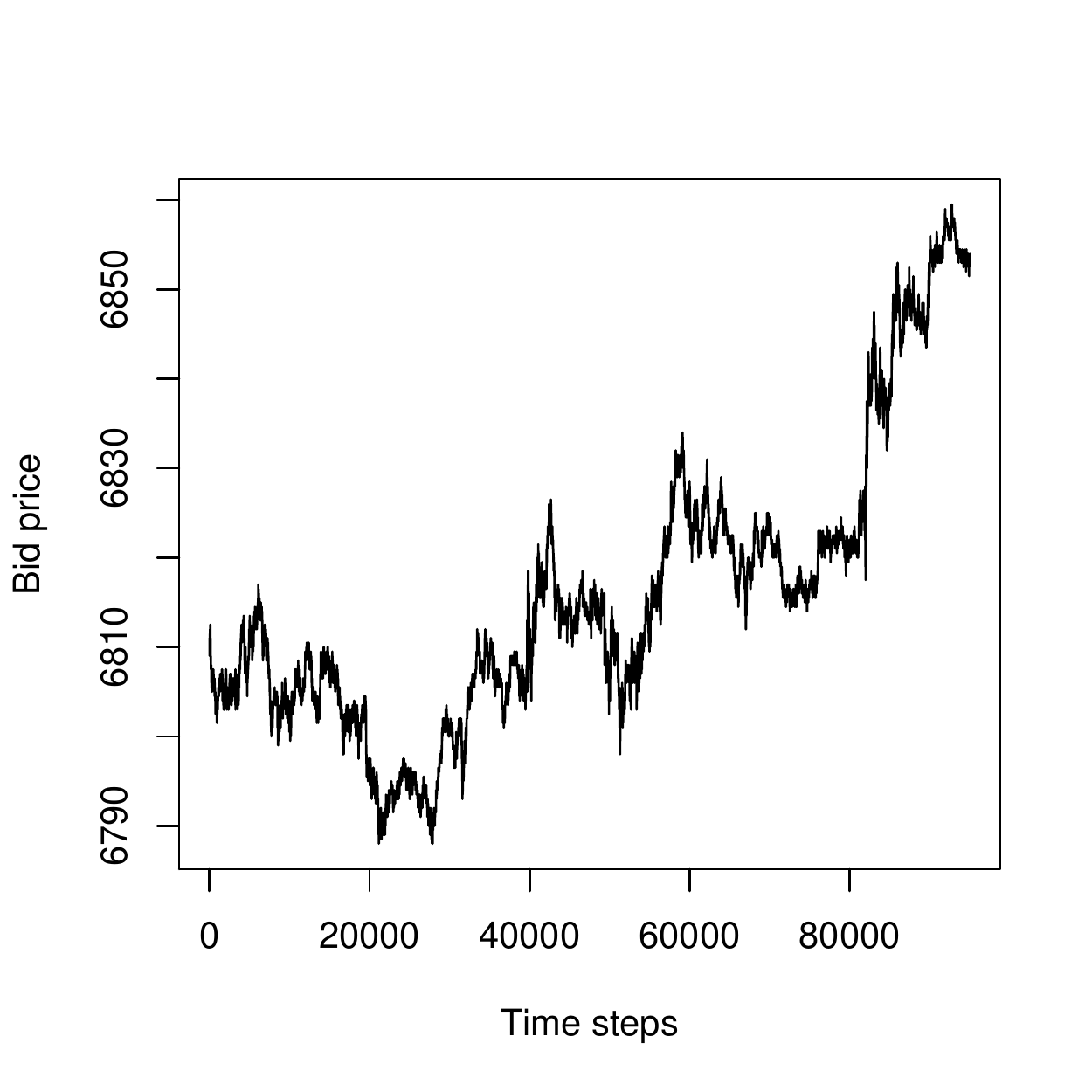}
\caption{The signal used: a commodity future, with millisecond time sampling.}
\label{fig100}
\end{figure*}

We use financial futures, circa March 2007, denominated in euros from the DAX
exchange.  Our  data stream is at the millisecond rate, and comprises about 
382,860
records.  Each record includes: 5 bid and 5 asking prices, together with
bid and asking sizes in all cases, and action.  We extracted one symbol
(commodity) with 95,011 single bid values, on which we now report results.
See Figure \ref{fig100}.

Embeddings were defined as follows.

\begin{itemize}
\item Windows of 100 successive values, starting at time steps:
1, 1000, 2000, 3000, 4000, $\dots$, 94000.
\item Windows of 1000 successive values, starting at time steps:
1, 1000, 2000, 3000, 4000, $\dots$, 94000.
\item Windows of 10000 successive values, starting at time steps:
1, 1000, 2000, 3000, 4000, $\dots$, 85000.
\end{itemize}

\begin{figure*}
\includegraphics[width=16cm]{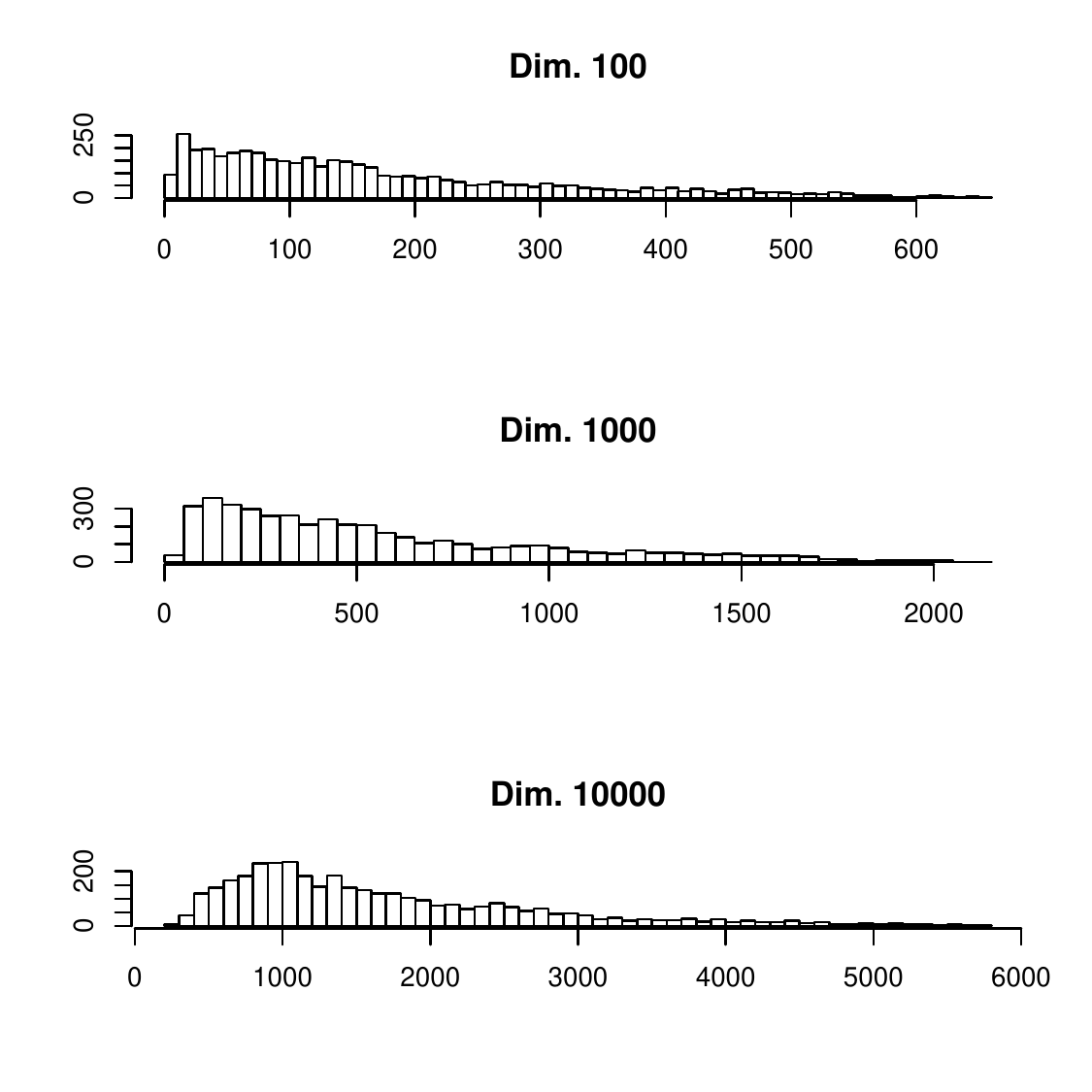}
\caption{Histograms of pairwise distances between
embeddings in dimensionalities 100, 1000, 10000.
Respectively the numbers of embeddings are: 95, 95 and 86.}
\label{fig110}
\end{figure*}

The histograms of distances between these windows, or embeddings, in 
respectively
spaces of dimension 100, 1000 and 10000, are shown in Figure \ref{fig110}.

Note how the
10000-length window case results in points that are strongly overlapping.
In fact, we can say that 90\% of the values in each window are overlapping
with
the next window.  Notwithstanding this major overlapping in regard to clusters
involved in the pairwise distances, if we can still find clusters in the
data then we have a very versatile way of tackling the clustering objective.
Because of the greater cluster concentration that we expect (cf.\
Table \ref{tabunifgauss}) from a greater embedding
dimension, we use the 86 points in 10000-dimensional space, notwithstanding
the fact that these points are from overlapping clusters.

We make the following supposition based on Figure \ref{fig100}: the
clusters will consist of successive values, and hence will be justifiably
termed segments.

From the distances
histogram in Figure \ref{fig110}, bottom, we will carry out Gaussian
mixture modeling followed by use of the Bayesian information criterion (BIC,
\cite{schwarz})
as an approximate Bayes factor, to determine the best number of
clusters (effectively, histogram peaks).  

\begin{figure*}
\includegraphics[width=8cm]{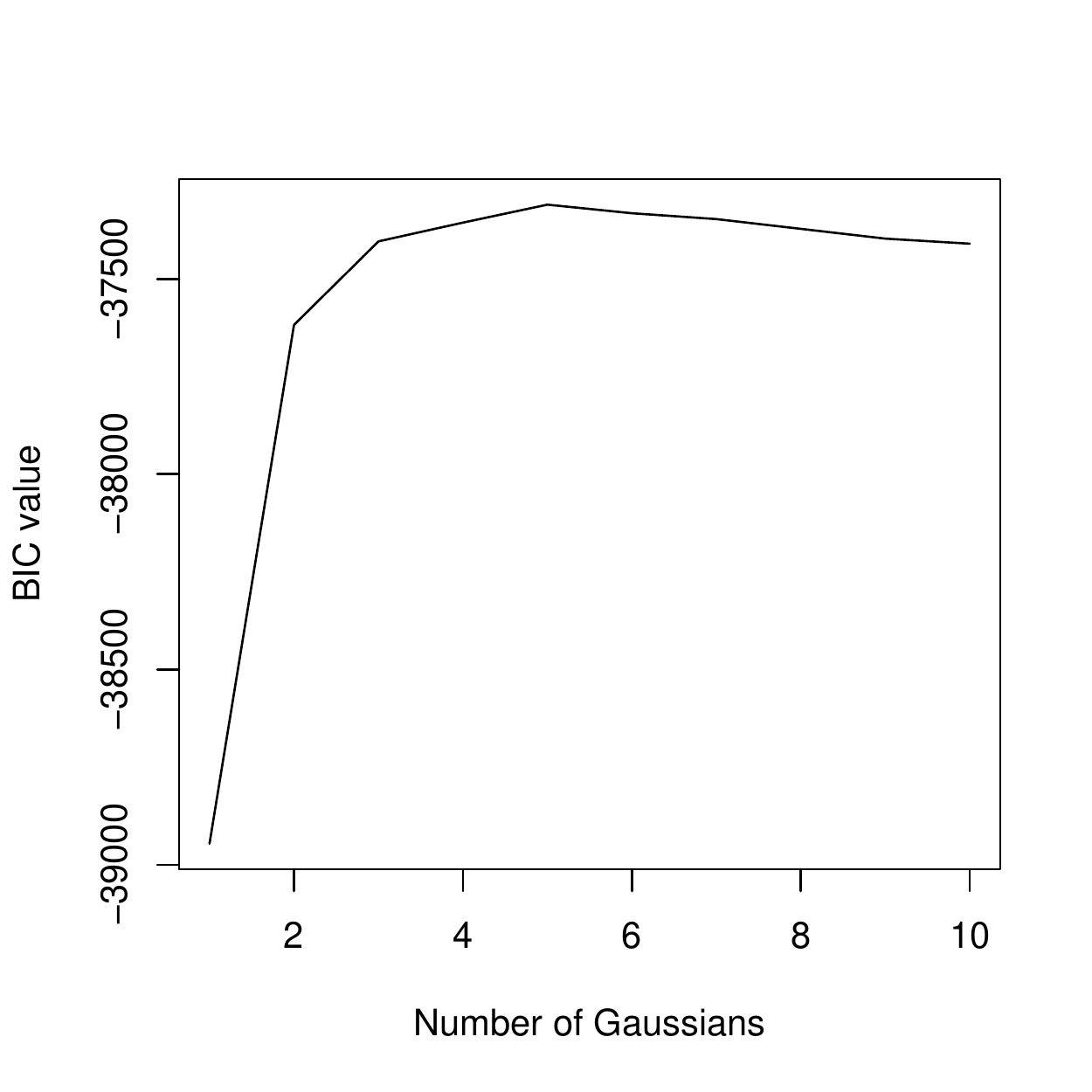}
\caption{BIC (Bayesian information criterion) values for the succession of
results.  The 5-cluster solution has the highest value for BIC and is therefore
the best Gaussian mixture fit.}
\label{fig120}
\end{figure*}

We fit a Gaussian mixture model
to the data shown in the bottom histogram of Figure \ref{fig110}.
To derive the appropriate number of histogram peaks we fit Gaussians and use
the Bayesian information criterion (BIC) as an approximate Bayes factor for
model selection \cite{kass,murtsta03}.
Figure \ref{fig120} shows the
succession of outcomes, and indicates as best a 5-Gaussian fit.
For this result, we find  the means of the Gaussians
to be as follows:  517, 885, 1374, 2273 and 3908.  The corresponding
standard deviations are: 84, 133, 212, 410 and 663.  The respective
cardinalities
of the 5 histogram peaks are: 358, 1010, 1026, 911 and 350.  Note that
this relates
so far only to the histogram of pairwise distances.  We now want to determine
the corresponding clusters in the input data.

While we have the segmentation of the distance histogram, we need the
segmentation of the original financial signal.  If we had 2 clusters in the
original financial signal, then we could expect up to 3 peaks in the
distances histogram (viz., 2 intra-cluster peaks, and 1 inter-cluster peak).
If we had 3 clusters in the original financial signal, then we could
expect up to 6 peaks in the distances histogram (viz., 3 intra-cluster
peaks, and 3 inter-cluster peaks).  This information is consistent
with asserting that the evidence from Figure \ref{fig120} points to
two of these histogram peaks being approximately co-located (alternatively:
the distances are approximately the same).  We conclude that 3 clusters
in the original financial signal is the most consistent number of clusters.
We will now determine these.

\begin{figure*}
\includegraphics[width=12cm]{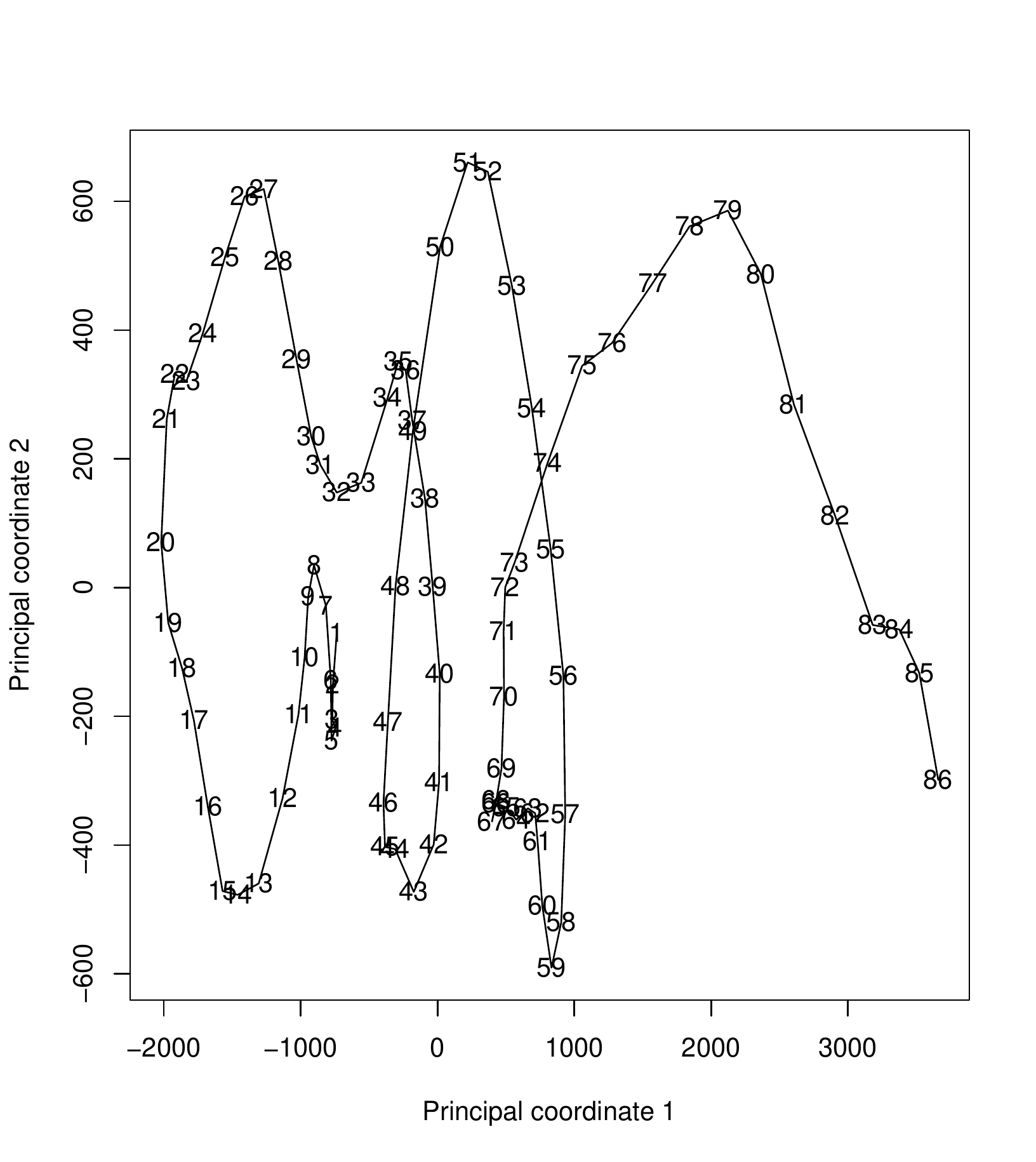}
\caption{An interesting representation -- a type of ``return map'' -- found
using a principal coordinates analysis of the 86 successive
 10000-dimensional points.
Again a demonstration that very high dimensional structures can be of
very simple structure.  The planar projection seen here represents most
of the information content of the data: the first axis accounts for
88.4\% of the variance, while the second accounts for 5.8\%.}
\label{fig180}
\end{figure*}

One possibility is to use principal coordinates analysis
(Torgerson's, Gower's metric multidimensional scaling) of the pairwise
distances.  In fact, a
2-dimensional mapping furnishes a very similar pairwise
distance histogram to that seen using
the full, 10000, dimensionality.  The first axis in Figure \ref{fig180}
accounts for 88.4\% of the variance, and the second for 5.8\%.
Note therefore how the scales of the planar representation in Figure
\ref{fig180} point to it being very linear.

Benz\'ecri (\cite{benz2}, chapter 7,
section 3.1) discusses the Guttman effect, or Guttman scale, where
factors that are not mutually correlated, are nonetheless
functionally related.  When there is a ``fundamentally unidimensional 
underlying phenomenon'' (there are multiple such cases here)
factors are functions of Legendre polynomials.
We can view Figure \ref{fig180} as consisting of multiple horseshoe shapes.
A simple explanation for such shapes is in terms of the
constraints imposed by lots of equal distances when the data vectors are
ordered linearly (see \cite{murtca}, pp.\ 46-47).

Another view of how embedded (hence clustered) data are capable of
being well mapped into a unidimensional curve is Critchley and Heiser
\cite{heiser1d}.  
Critchley and Heiser show one approach to mapping an ultrametric into
a linearly or totally ordered metric.  We have asserted and then established
how hierarchy in some form is relevant for high dimensional data spaces; and
then we find a very linear projection in Figure \ref{fig180}.  As a consequence
we note that the Critchley and Heiser result is especially relevant for
high dimensional data analysis.

Knowing that 3 clusters in the original signal are wanted,
we could use Figure \ref{fig180}.  There are various ways to do 
so.

\begin{figure*}
\includegraphics[width=20cm,angle=270]{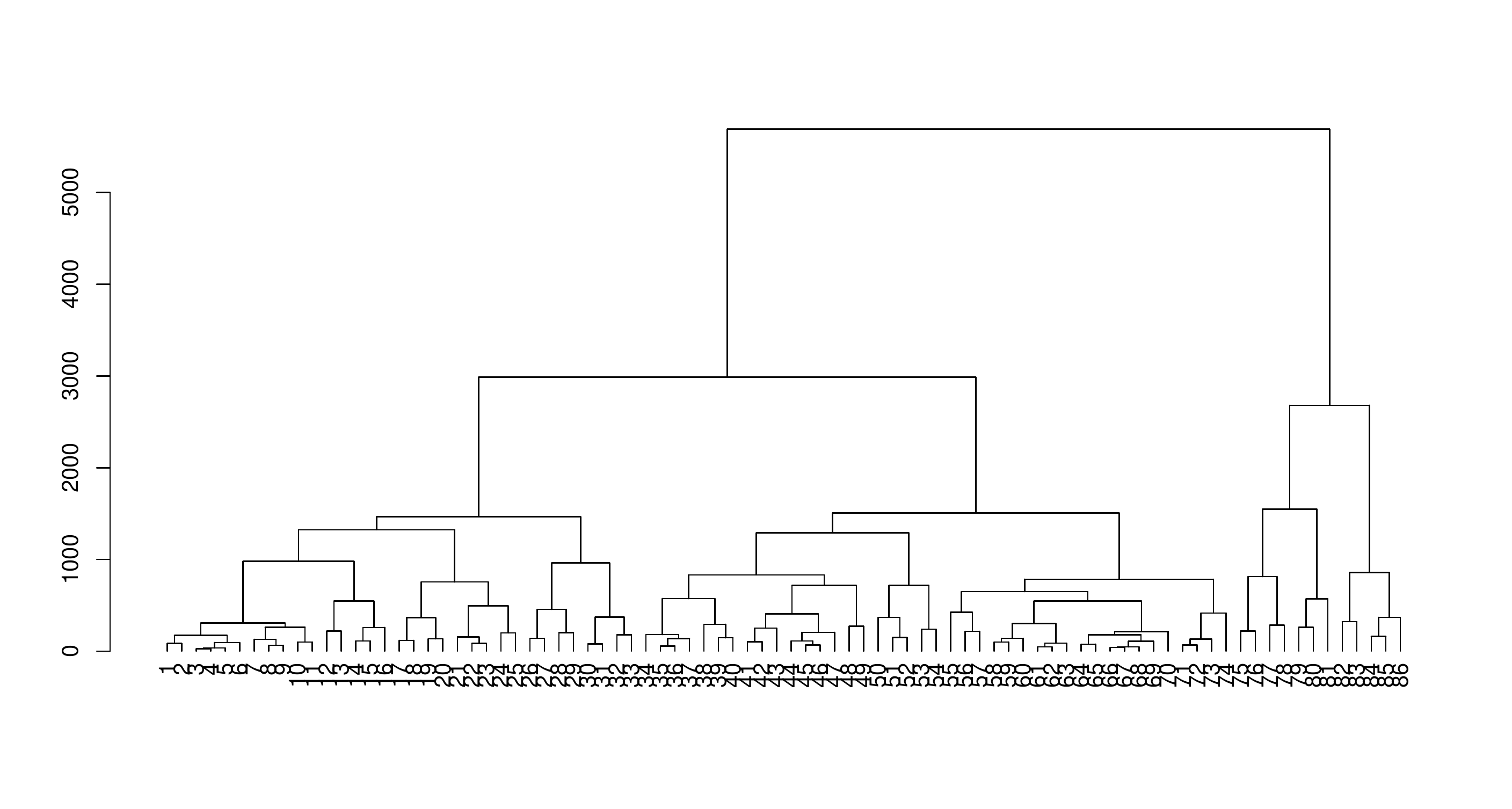}
\caption{Hierarchical clustering of the 86 points.
Sequence is respected.  The agglomerative criterion is the
contiguity-constrained complete link method.  See \cite{murt85} for
details including proof that there can be no inversion in this dendrogram.}
\label{fig140}
\end{figure*}

We will use an adjacency-constrained agglomerative
hierarchical clustering algorithm to find the clusters:
see Figure \ref{fig140}.  The contiguity-constrained
complete link criterion
is our only choice here if we are to be sure that no inversions
can come about in the hierarchy, as explained in \cite{murt85}.
As input, we use the coordinates in Figure \ref{fig180}.
The 2-dimensional Figure \ref{fig180} representation relates
to over 94\% of the variance.  The most complete basis was of
dimensionality 85.  We checked the results of the 85-dimensionality
embedding which, as noted below, gave very similar results.

\begin{figure*}
\includegraphics[width=16cm]{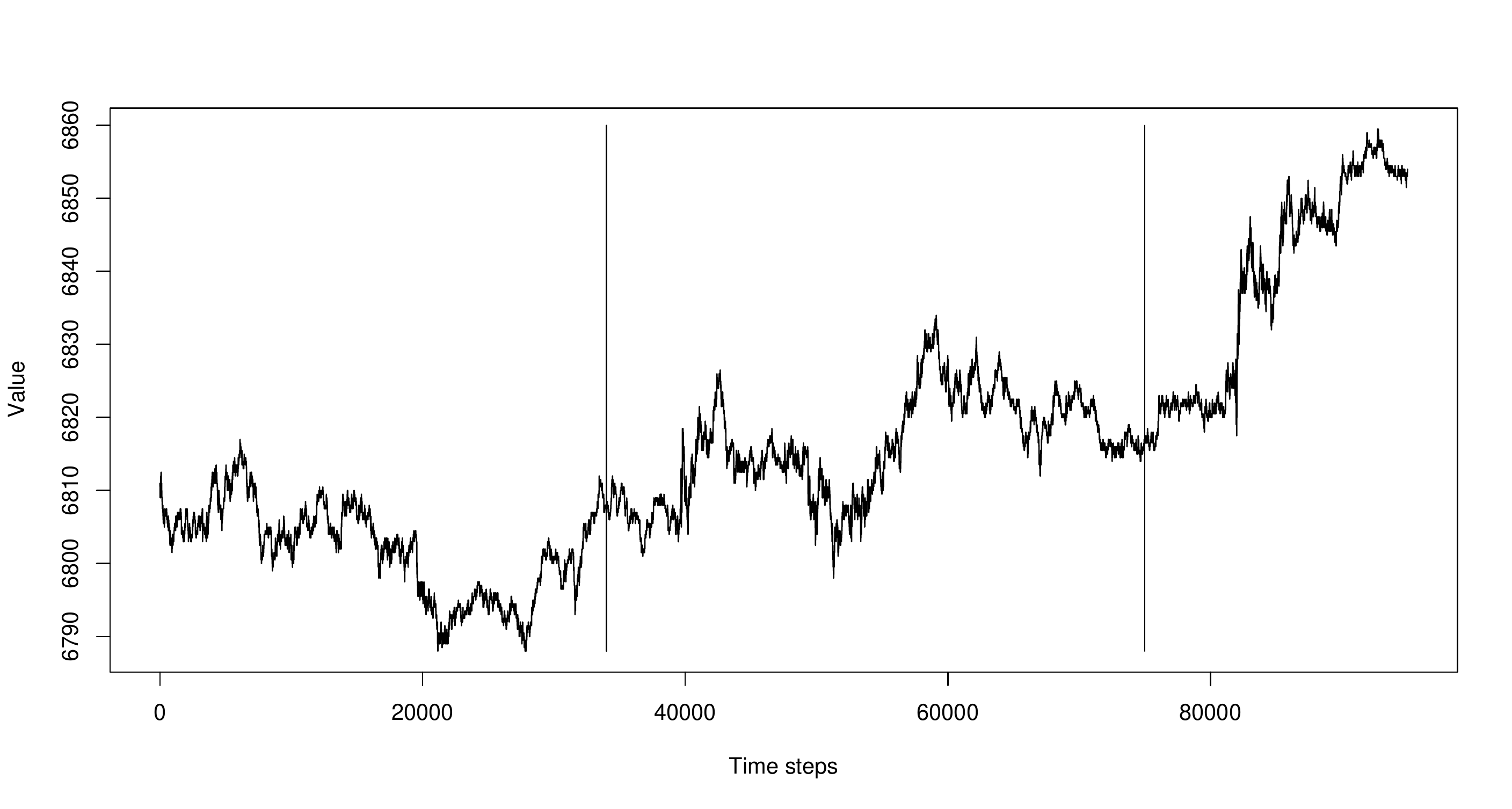}
\caption{Boundaries found for 3 segments.}
\label{fig160}
\end{figure*}

Reading off the 3-cluster memberships from Figure \ref{fig140}
gives for the signal actually used (with a very initial segment
and a very final segment deleted): cluster 1 corresponds to signal
values 1000 to 33999 (points 1 to 33 in Figure \ref{fig140});
cluster 2 corresponds to signal values 34000 to 74999 (points 34 to 74
in Figure \ref{fig140});
and cluster 3 corresponds to signal values 75000 to 86999 (points
75 to 86 in Figure \ref{fig140}).
This allows us to segment the original time series: see Figure \ref{fig160}.
(The clustering of the 85-dimensional embedding differs minimally.
Segments are: points 1 to 32; 33 to 73; and 74 to 86.)

To summarize what has been done:

\begin{enumerate}
\item the segmentation is initially guided by the peak-finding in the
histogram of distances
\item with high dimensionality we expect simple structure in a low
dimensional mapping provided by principal coordinates analysis
\item either the original high dimensional data or the principal
coordinates analysis embedding are used
as input to a sequence-constrained clustering
method in order to determine the clusters
\item which can then be displayed on the original data.
\end{enumerate}

In this case, the clusters are defined using a complete link criterion,
implying that these three clusters
are determined by minimizing their maximum internal pairwise distance.
This provides a strong measure of signal volatility as an
explanation for the clusters, in addition to their average value.

\section{Conclusions}

Among themes not covered in this article are data stream clustering.
To provide background and motivaton, in \cite{steklov}, we discuss 
permutation representations of a data stream.  Since hierarchies
can also be represented as permutations, there is a ready way to 
associate data streams with hierarchies.  In fact, early computational
work on hierarchical clustering used permutation representation to 
great effect (cf.\ \cite{sibson}).  
To analyze data streams in this way, in 
\cite{murtaghEPJ} we develop an approach to ultrametric embedding of
time-varying signals, including biomedical, meteorological, financial
and other.  This work has been pursued in physics by Khrennikov.  

Let us now wrap up on the exciting perspectives opened up by our 
work on the theme of symmetry-finding through hierarchy in very large
data collections.  

``My thesis has been that one path to the construction of a nontrivial
theory of complex systems is by way of a theory of hierarchy.'' 
Thus Simon (\cite{simon}, p. 216).   
We have noted symmetry in many guises in the representations used, in the
transformations applied, and in the transformed outputs.  These symmetries
are non-trivial too, in a way that would not be the case were we 
simply to look at classes of a partition and claim that cluster members were
mutually similar in some way.    We have seen how the p-adic or ultrametric
framework provides significant focus and commonality of viewpoint.  

Furthermore we have highlighted the computational scaling properties
of our algorithms.  They are fully capable of addressing the data and
information deluge that we face, and providing us with the best 
interpretative and decision-making tools.  The full elaboration of this
last point is to sought in each and every application domain, and 
face to face with old and new problems.  

In seeking (in a general way) and in determining (in a focused way) 
structure and regularity in massive data stores, we see that,
in line with the insights and achievements of Klein, Weyl and Wigner, 
in data mining and data analysis we seek and determine symmetries in 
the data that express observed and measured reality.  

\bibliographystyle{plain}
\bibliography{data-mining-murtagh}

\end{document}